\documentclass[10pt,twocolumn,letterpaper]{article}

\usepackage{iccv}
\usepackage{times}
\usepackage{epsfig}
\usepackage{graphicx}
\usepackage{amsmath}
\usepackage{amssymb}

\usepackage[pagebackref=true,breaklinks=true,letterpaper=true,colorlinks,bookmarks=false]{hyperref}

\def\iccvPaperID{7141} 

\usepackage{xcolor}
\usepackage{mathtools}
\usepackage{soul}

\newif\ifdraft
\draftfalse

\ifdraft
\newcommand{\dcc}[1]{{\color{red}[\textbf{Danny:} #1]}}
\newcommand{\kac}[1]{{\color{blue}[\textbf{Kfir:} #1]}}
\newcommand{\avc}[1]{{\color{purple}[\textbf{Andrey:} #1]}}
\newcommand{\cqc}[1]{{\color{violet}[\textbf{Chu:} #1]}}

\newcommand{\ka}[1]{{\color{blue}#1}}

\newcommand{\cq}[1]{{\color{violet}#1}}

\newcommand{\drop}[1]{}

\else
\newcommand{\dcc}[1]{}
\newcommand{\kac}[1]{}
\newcommand{\avc}[1]{}
\newcommand{\cqc}[1]{}

\newcommand{\ka}[1]{{\color{black}#1}}

\newcommand{\cq}[1]{{\color{black}#1}}
\fi

\newcommand{\Prompt}{\mathcal{P}}
\newcommand{\PromptP}{\mathcal{P}+}

\makeatletter
\DeclareRobustCommand\onedot{\futurelet\@let@token\@onedot}
\def\@onedot{\ifx\@let@token.\else.\null\fi\xspace}
\DeclareMathAlphabet\mathbfcal{OMS}{cmsy}{b}{n}

\makeatother

\makeatletter
\def\blfootnote{\xdef\@thefnmark{}\@footnotetext}
\makeatother

\newif\ifwatermark
\watermarktrue
\draftfalse

\newcommand*{\affaddr}[1]{#1} 
\newcommand*{\affmark}[1][*]{\textsuperscript{#1}}

\usepackage[symbol]{footmisc}

\begin{document}

\title{$\mathcal{P}+$: Extended Textual Conditioning in Text-to-Image Generation}

\newcommand{\ourmethod}{$\mathcal{P}+$}

\author{%
Andrey Voynov \affmark[1], Qinghao Chu\affmark[1], Daniel Cohen-Or\footref{note}~~\affmark[1,]\affmark[2], Kfir Aberman\affmark[1]\\
\small{\affaddr{\affmark[1]Google Research,~~}}\small{\affaddr{\affmark[2]The Blavatnik School of Computer Science, Tel Aviv University}}\
}

\maketitle


\begin{abstract}
    We introduce an Extended Textual Conditioning space in text-to-image models, referred to as $\PromptP$. This space consists of multiple textual conditions, derived from per-layer prompts, each corresponding to a layer of the denoising U-net of the diffusion model. 
We show that the extended space provides greater disentangling and control over image synthesis. We further introduce Extended Textual Inversion (XTI), where the images are inverted into $\PromptP$, and represented by per-layer tokens.
We show that XTI
is more expressive and precise, and converges faster than the original Textual Inversion (TI) space.
The extended inversion method does not involve any noticeable trade-off between reconstruction and editability and induces more regular inversions.
We conduct a series of extensive experiments to analyze and understand the properties of the new space, and to showcase the effectiveness of our method for personalizing text-to-image models. Furthermore, we utilize the unique properties of this space to achieve previously unattainable results in object-style mixing using text-to-image models.
\end{abstract}
\footnotetext[2]{\label{note}Performed this work while working at Google.}

\section{Introduction}

Neural generative models have advanced the field of image synthesis, allowing us to create incredibly expressive
and diverse images. 
Yet, recent breakthroughs in text-to-image models based on large language-image models have taken this field to new heights and stunned us with their ability to generate images from textual descriptions, providing a powerful tool for creative expression, visualization, and design.

\begin{figure}[h]
    \centering
    \includegraphics[width=0.49\columnwidth]{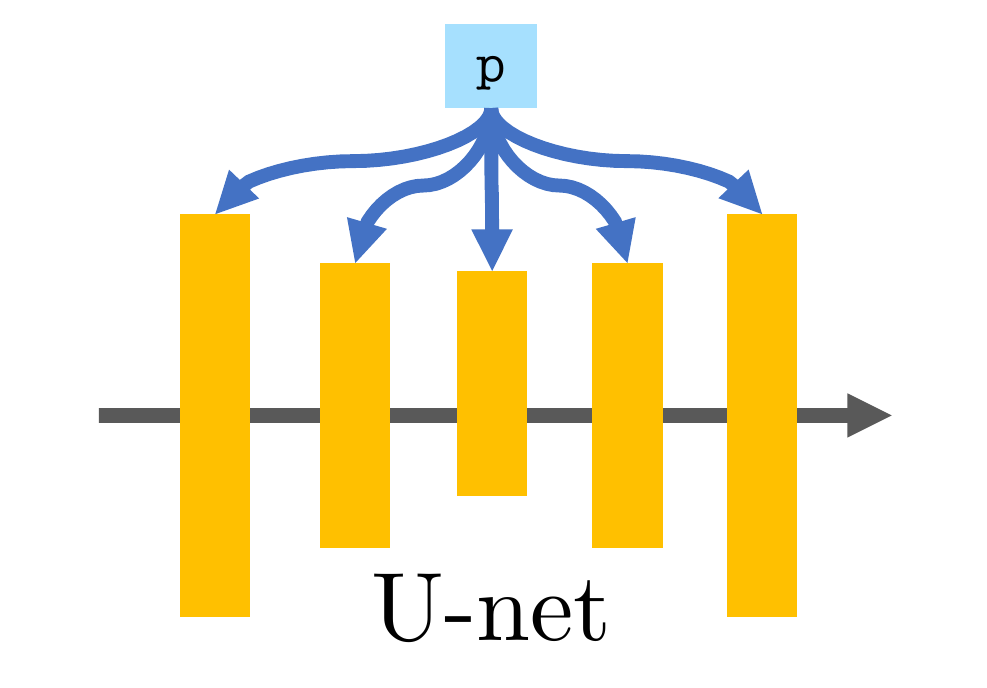}
    \includegraphics[width=0.49\columnwidth]{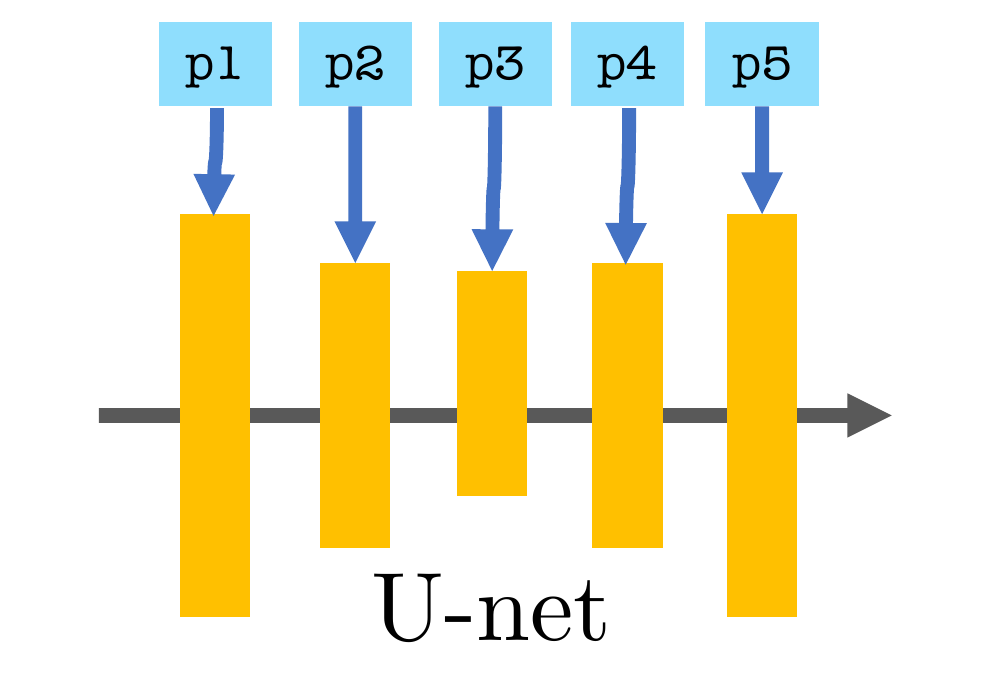}
    \caption{{\bf $\mathbfcal{P}$ vs. $\mathbfcal{P}+$}. Standard textual conditioning, where a single text embedding is injected to the network (left), vs. our proposed extended conditioning, where different embeddings are injected into different layers of the U-net (right).}
    \label{fig:unet}
    \vspace{-10pt}
\end{figure}

\begin{figure}[b]
    \centering
    \includegraphics[width=\columnwidth]{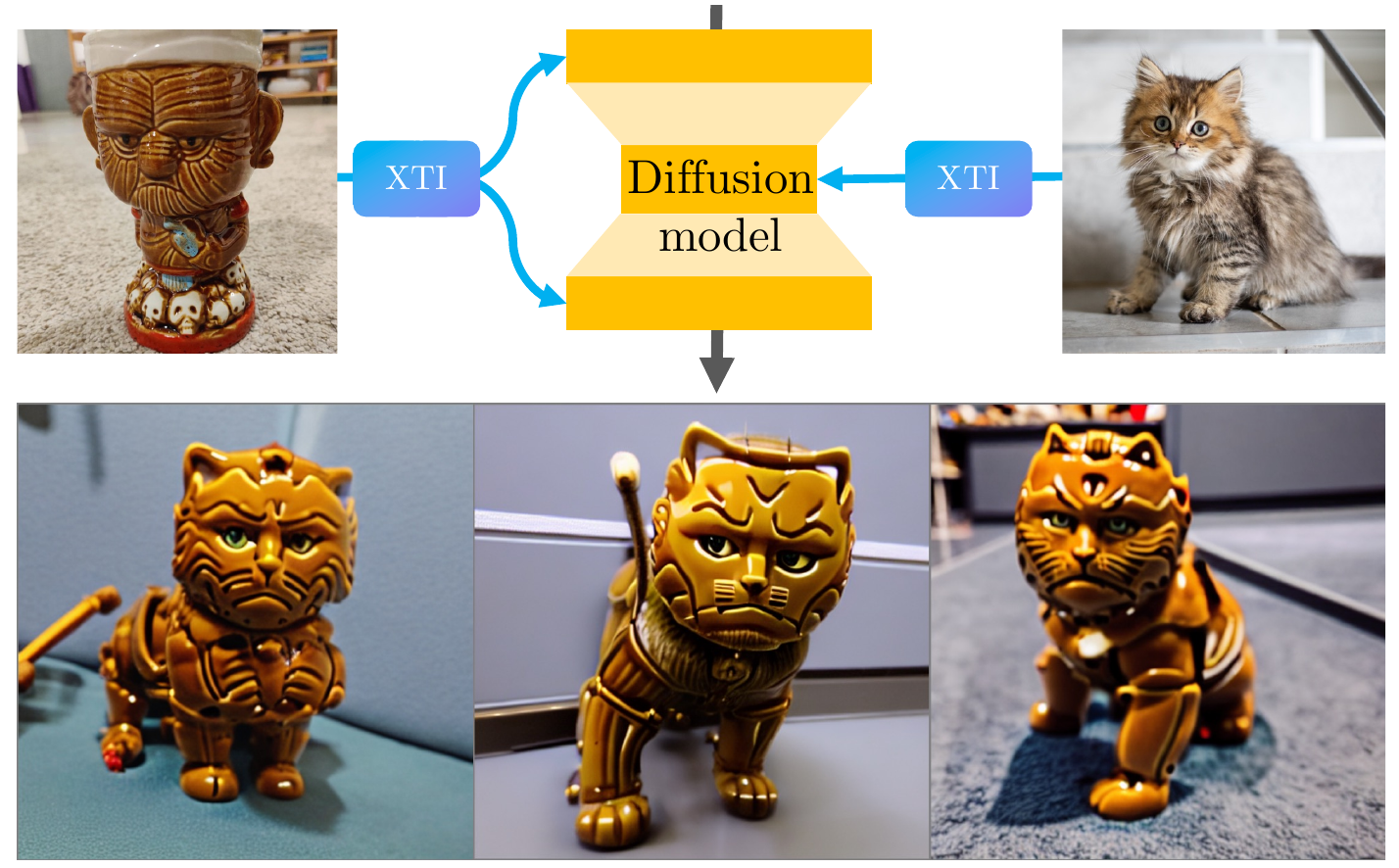}
    \caption{
    {\bf Shape-Style Mixing in XTI. }The extended textual space allows mixing concepts learned from two separate extended textual inversions (XTIs). The inversion of the kitten (right) is injected to the coarse inner layers of the U-net, affecting the shape of the generated image, and the inversion of the cup (left) is injected to the outer layers, affecting the style and appearance.  
    }
    \label{fig:teaser_mix}
\end{figure}

These text-to-image diffusion models use the encoded text as conditioning. We can refer to the conditioning space defined by the tokens embedding space of the language model as $\Prompt$ space. 
In other words, $\Prompt$ is the \textit{textual-conditioning space}, where during synthesis, an instance $p \in \Prompt$ (after passing through a text encoder) is injected to all attention layers of a U-net, as illustrated in Figure \ref{fig:unet} (left).
In this paper, we introduce the \textit{Extended Textual Conditioning} space. This space, referred to as $\PromptP$ space, consists of $n$ textual conditions $\{p_1, p_2, ... p_n\}$, where each $p_i$ is injected to the corresponding layer $i$ in the U-net (see Figure \ref{fig:unet} (right)). $\mathcal{P}+$ space is more expressive, disentangled and thus provides better control on the synthesized image. As will be analyzed in this paper, different layers have varying degrees of control over the attributes of the synthesized image. In particular, the coarse layers primarily affect the structure of the image, while the fine layers predominantly influence its appearance.


The introduction of $\mathcal{P+}$ textual conditioning space opens the door to a particularly exciting advancement in the domain personalization of text-to-image models~\cite{gal2022image,ruiz2022dreambooth}, where the model learns to represent a specific concept described in a few input images as a dedicated token. This learned token can then be employed in a text prompt to produce diverse and novel images related to the concept provided by the user. This technique of learning tokens is referred to as Textual Inversion (TI) \cite{gal2022image}.

In our work, we introduce Extended Textual Inversion (XTI), where we invert the input images into a set of token embeddings, one per layer, namely, inversion into $\PromptP$.
Our findings reveal that the expanded inversion process in $\PromptP$ is not only faster than TI, but also more expressive and precise, owing to the increased number of tokens that provide superior reconstruction capabilities. Remarkably, the improved reconstruction does not compromise editability, as demonstrated by our results. 

Furthermore, we leverage the distinctive characteristics of $\PromptP$ to advance the state-of-the-art in object-appearance mixing through text-to-image generation. Specifically, we employ the insertion of inverted tokens of diverse subjects into the different layers to capitalize on the inherent shape-style disentanglement exhibited by these layers. This approach enables us to achieve previously unattainable results as shown in Figure~\ref{fig:teaser_mix}.

We conduct extensive experiments and evaluation to demonstrate the effectiveness of the new space, analyzing its properties and showcasing its power for personalizing text-to-image models, object-style mixing, and more. Project page: \url{https://prompt-plus.github.io}
\section{Related works}

\subsection{Extended 
Spaces in Generative Models}

Exploring neural sub-spaces in generative models has been extensively explored, most notably in
StyleGAN~\cite{Karras2020ada,karras2019style}.
The extended textual conditioning $\PromptP$ is reminiscent of StyleGAN's extended latent space \cite{abdal2019image2stylegan,abdal2020image2stylegan++}, also commonly referred to as $\mathcal{W+}$. 
Similar to $\mathcal{W+}$, $\mathcal{P+}$ is significantly more expressive, where instead of a single code shared by all layers, there is one per layer. However, while $\mathcal{W+}$ is an extended latent space, here the extended space relates to the textual conditions used by the network. It should be noted, though, that while $\mathcal{W+}$ is expressive, the extended code is less editable \cite{tov2021designing}. In contrast, $\mathcal{P+}$ remains practically as editable as $\mathcal{P}$. In addition, other sub-spaces lay within deeper and more disentangled layers \cite{wu2021stylespace} have been explored and exploited in various editing and synthesis applications \cite{bermano2022state}.


In the case of text-to-image diffusion models, the denoising U-net, which is the core model of most of the text-to-image diffusion models, is usually conditioned by text prompts via a set of cross-attention layers \cite{ramesh2022hierarchical, rombach2021highresolution, saharia2022photorealistic}. 
In many neural architectures, different layers are responsible for different abstraction levels \cite{bau2020units, karras2019style, voynovrpgan, zeiler2014visualizing}. It is natural to anticipate that the diffusion denoising U-Net backbone operates in a similar manner, with different textual descriptions and attributes proving beneficial at different layers.

\subsection{Text-Driven Editing}

There has been a significant advancement recently in generating images based on textual inputs through Text-to-Image models~\cite{chang2023muse, ramesh2022hierarchical,rombach2021highresolution, saharia2022photorealistic}, where most of them exploit the powerful architecture of diffusion models \cite{ho2020denoising,rombach2021highresolution,sohl2015deep,song2020denoising,song2019generative}.

In particular, recent works have attempted to adapt text-guided diffusion models to the fundamental problem of single-image editing, aiming to exploit their rich and diverse semantic knowledge of this generative prior.

In a pioneering attempt, Meng et al.~\cite{meng2021sdedit} add noise to the input image and then perform a denoising process from a predefined step. Yet, they struggle to accurately preserve the input image details, which were preserved by a user provided mask in other works~\cite{avrahami2022blended, avrahami2022blendedlatent, nichol2021glide}. DiffEdit \cite{couairon2022diffedit} employs DDIM inversion for image editing, but to prevent any resulting distortion, it generates a mask automatically that allows background preservation.

Text-only editing approaches split into approach that supports to global editing  \cite{crowson2022vqgan, kim2021diffusionclip, kwon2021clipstyler, patashnik2021styleclip}, and local editing ~\cite{bar2022text2live,wang2022imagen}. Prompt-to-prompt~\cite{hertz2022prompt} introduces an intuitive editing technique that enables manipulation of local or global details by injecting internal cross-attention maps.
To allow prompt-to-prompt to be applied to real images, Null-Text Inversion \cite{mokady2022null} is proposed as means to invert real images into the latent space of the diffusion model.
Imagic \cite{Kawar2022ImagicTR} and UniTune~\cite{valevski2022unitune} have demonstrated impressive text-driven editing capabilities, but require the costly fine-tuning of the model. The InstructPix2Pix~\cite{brooks2022instructpix2pix}, Plug-and-Play~\cite{pnpDiffusion2022}, and Parmar et al.~\cite{parmar2023zero} allow users to input an instruction or target prompt and manipulate real images accordingly, to achieve the desired edits.

 

\subsection{Personalization}
Synthesizing particular concepts or subjects which are not widespread in the training data is a challenging task. 
This requires an \textit{inversion} process that given input images would enable regenerating the depicted object using a text-guided diffusion model. Inversion has been studied extensively for GANs \cite{bermano2022state, creswell2018inverting, lipton2017precise, xia2021gan, yeh2017semantic, zhu2016generative}, ranging from latent-based optimization \cite{abdal2019image2stylegan,abdal2020image2stylegan++} and encoders \cite{richardson2020encoding,tov2021designing} to feature space encoders \cite{Wang2021HighFidelityGI} and fine-tuning of the model \cite{alaluf2021hyperstyle, roich2021pivotal, nitzan2022mystyle}. 

The notion of personalization of text-to-image models has been shown to be a powerful technique. 
Personalization of models \cite{kumari2022customdiffusion, ruiz2022dreambooth} in general or of text tokens \cite{gal2022image} has quickly been adapted for various applications \cite{kawar2022imagic,lin2022magic3d}. In addition to their high computational cost, current methods face a clear-trade-off between learning tokens that accurately capture concepts vs. avoidance of overfitting. This can result in learned tokens that are overly tuned to the input images, thus limiting their ability to generalize to new contexts or generate novel variations of the concept. 

Similar to TI, our approach does not require any fine-tuning or modification of the weights, thus, reduces the risk of overfitting and degrading the editability capabilites. In contrast, our inversion process into $\PromptP$ is both faster and more precise, thanks to the greater number of tokens that improve reconstruction capabilities without sacrificing editability. 




\section{Extended Conditioning Space}


\begin{figure}
    \centering
    \includegraphics[width=\columnwidth]{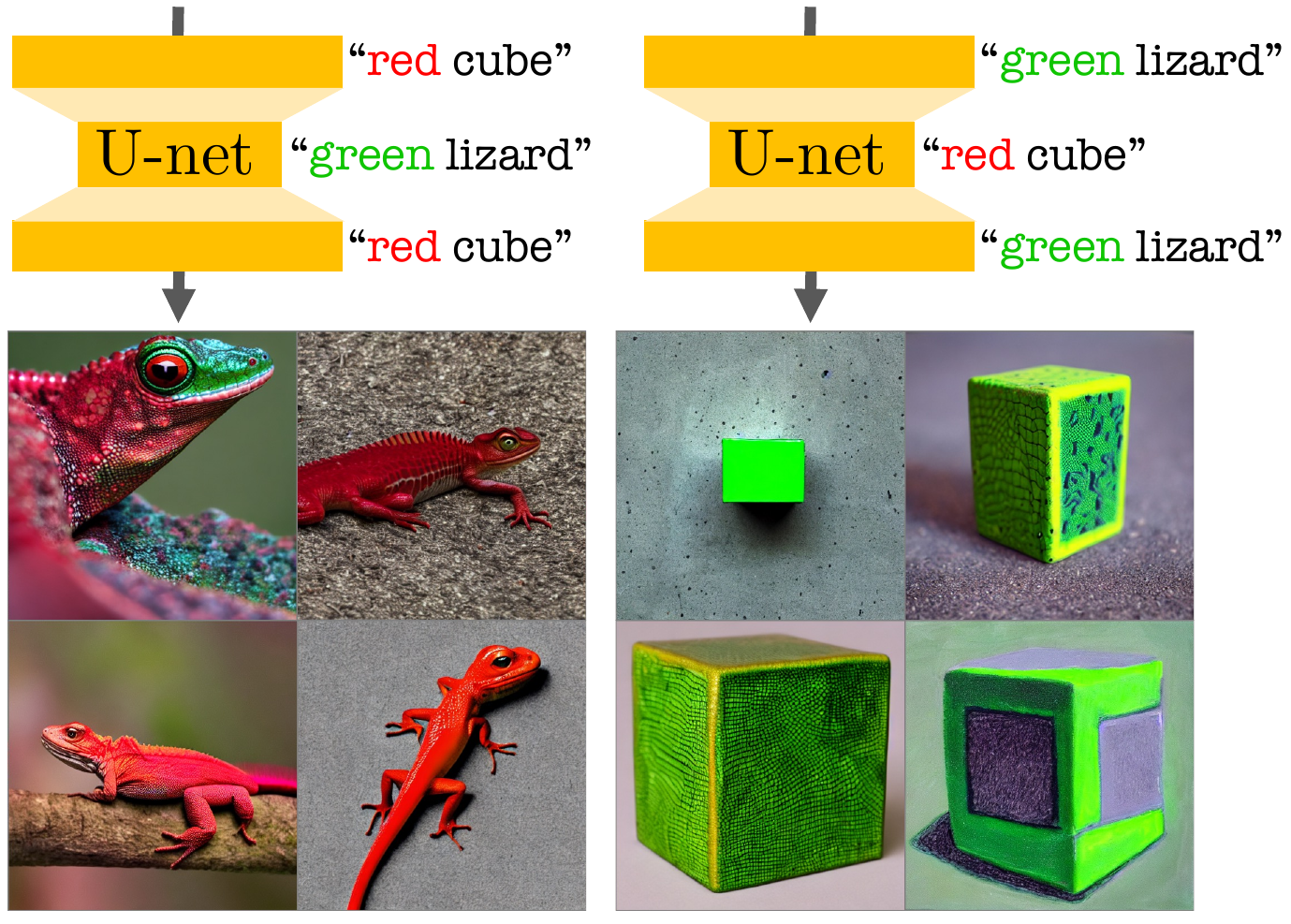}
    \caption{{\bf Per-layer Prompting.} We provide different text prompts (a precursor to $\PromptP$) to different cross-attention layers in the denoising U-net. We see that color (\texttt{"red"}, \texttt{"green"}) is determined by the fine outer layers and content (\texttt{"cube"}, \texttt{"lizard"}) is determined by the coarse inner layers.
    }
    \label{fig:swap}
    \vspace{-10pt}
\end{figure}

To engage the reader, we begin with a simple experiment on the publicly available Stable Diffusion model \cite{Rombach_2022_CVPR}.
We partitioned the cross-attention layers of the denoising U-net into two subsets: coarse layers with low spatial resolution and fine layers with high spatial resolution. We then used two conditioning prompts: \texttt{"red cube"} and \texttt{"green lizard"}, and injected one prompt into one subset of cross-attention layers, while injecting the second prompt into the other subset.
The resulting generated images are provided in Figure \ref{fig:swap}. Notably, at the first run the model generates a red lizard, by taking the subject from the coarse layers' text conditioning, and appearance from the fine layers' conditioning. Similarly, in the second run it generates the green cube, once again taking the appearance from the fine layers and the subject from the coarse layers. This experiment suggests that the conditioning mechanism at different resolutions processes prompts differently, with different attributes exerting greater influence at different levels. With this in mind, our work aims to further explore this phenomena and its potential applications.


In the following parts, we introduce the Extended Textual Conditioning space (\ourmethod) and its key properties. We then detail how this space can be utilized to perform textual inversion for a given set of images.

\subsection{$\mathbfcal{P}+$ Definition}


Let $\Prompt$ denote the \textit{textual-conditioning space}. $\Prompt$ refers to the space of token embeddings that are passed into the text encoder in a text-to-image diffusion model. To clarify the definition of this space, we provide a brief overview of the process that a given text prompt undergoes in the model before being injected into the denoising network.

\begin{figure}[h]
    \centering
    \includegraphics[width=0.9\columnwidth]{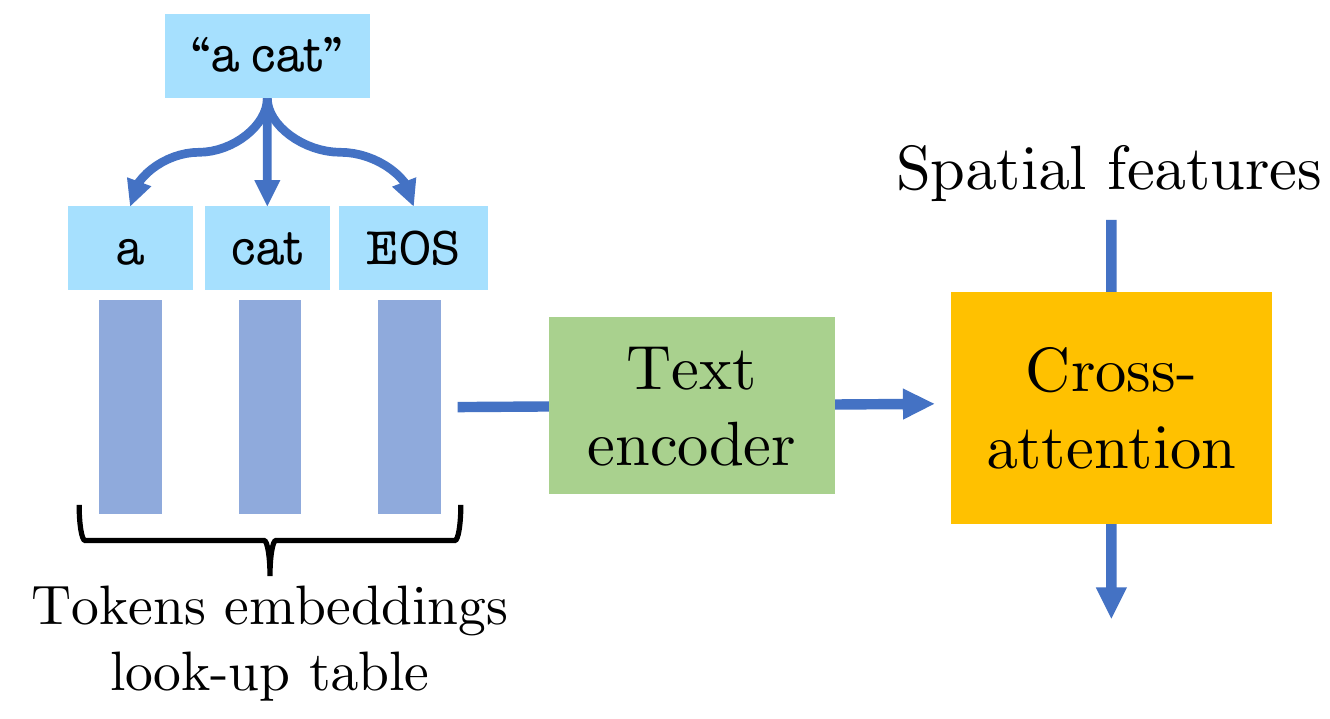}
    \caption{{\bf Text-conditioning mechanism of a denoising diffusion model.} The prompt \texttt{"a cat"} is processed with a sentence tokenization, by a pretrained textual encoder, and fed into a cross-attention layer. Each of the three bars on the left represents a token embedding in $\Prompt$.
    }
    \label{fig:cross_attention} 
\end{figure}

Initially, the text tokenizer splits an input sentence into tokens, with a special token marking the end of the sentence (EOS). Each token corresponds to a pre-trained embedding that is retrieved from the embedding lookup table. Subsequently, these embeddings are concatenated and passed through a pre-trained text encoder, then injected to the cross attention layers of the U-net model. In our work, we define $\Prompt$ as the set of individual token embeddings that are passed to the text encoder. The process of injecting a text prompt into the network for a particular cross-attention layer is illustrated in Figure~\ref{fig:cross_attention}.

We next present the \textit{Extended Textual Conditioning} space, denoted by $\PromptP$, which is defined as follows:
\begin{equation}
\PromptP  \coloneqq \left\{p_1, p_2, ... p_n\right\},
\end{equation}
where $p_i\in\Prompt$ represents an individual token embedding corresponding to the $i$-th cross-attention layer in the denoising U-net. Figure~\ref{fig:unet} illustrates the conceptual difference between the two spaces, $\Prompt$ (left) and $\PromptP$ (right).

With the definition of the new space, our diffusion model, previously conditioned on a single prompt $U(x | t, p)$, can now synthesize images in the extended space $U(x | t, {p_1, \dots, p_n})$, where $t$ denotes the denoising timestep.

As $\Prompt$ is a subspace of $\PromptP$, we naturally inquire about the advantages of synthesizing in the extended space. In Section~\ref{sec:exp}, we present an analysis of the properties of the new space, which showcases a higher degree of control over various attributes. Specifically, different layers are found to dominate different attributes, such as style, color, and structure.

A notable benefit of this space is its potential for enhancing textual inversion. We next demonstrate how the extended space can be utilized to represent subjects with greater fidelity, while maintaining the capability for editing.


\subsection{Extended Textual Inversion (XTI)}

Given a set of images $\mathcal{I} = \{I_1, \dots, I_k\}$ of a specific subject, the goal of the Textual Inversion (TI) operation \cite{gal2022image} is to find a representation of the object in the conditioning space $\Prompt$. We next explain how we extend the TI approach and perform the inversion into \ourmethod. This process is coined Extended Textual Inversion (XTI). 

First, we add $n$ new textual tokens $\texttt{t}_1, \dots, \texttt{t}_n$ to the tokenizer model, associated with $n$ new token embeddings lookup-table elements $e_1, \dots, e_n$. Then, similarly to \cite{gal2022image}, we optimize the token embeddings with the objective to predict the noise of a noisy images from $\mathcal{I}$, while the token embeddings are injected to the network. 

Assuming that the denoising U-net is parameterized by a set of parameters denoted by $\theta$, and operates within the extended conditioning space as previously described, we define the reconstruction objective for the embeddings $e_1, \dots, e_n$ that correspond to the tokens $\texttt{t}_1, \dots, \texttt{t}_n$ as follows:

\[ \mathcal{L}_{\mathrm{XTI}} = \mathop{\mathbb{E}}_{\begin{subarray} _P\sim \Pi,\ I \sim \mathcal{I},\\ \varepsilon \sim \mathcal{N}(0, 1),\ t\end{subarray}} \|\varepsilon - \varepsilon_\theta(I_t | t, P(\texttt{t}_1, \dots, \texttt{t}_n))\|_2^2 \]
where $I_t$ is the image $I$ noised with the additive noise $\varepsilon$ according to the noise level $t$. Once we operate with a latent diffusion model, we always suppose that $I$ is a latent image representation. The new look-up table embeddings $e_1, \dots, e_n$ that correspond to $\texttt{t}_1, \dots, \texttt{t}_n$ are optimized w.r.t. $\mathcal{L}_\mathrm{XTI}$.
This optimization is applied independently to each cross-attention layer.
\section{Experiments and Evaluation}
\label{sec:exp}

In this section, we conduct an in-depth analysis of the various properties exhibited by the U-net cross-attention layers, and investigate how these characteristics are distributed across the layers. This analysis constitutes a motivation for the effectiveness of our $\PromptP$ space. We then present a comprehensive evaluation of our proposed XTI approach for the personalization task, encompassing quantitative, qualitative, and user study analysis. For more details about the user study setting please refer to the supplementary material.


In all of our experiments we use the Stable Diffusion 1.4 model~\cite{Rombach_2022_CVPR}. It is built on top of CLIP~\cite{radford2021learning}, whose token embedding is represented by a vector with $768$ entries, such that $\Prompt\subseteq\mathbb{R}^{768}$. Stable Diffusion a latent diffusion model whose denoising U-net operates on an autoencoded image latent space. The U-net has four spatial resolution levels - 8x8, 16x16, 32x32, and 64x64. The 16, 32, and 64 resolution levels each have two cross-attention layers on the downward (contracting) path and three cross-attention layers on the upward (expansive) path. Resolution 8 has only 1 cross-attention layer. Thus there are a total of 16 cross-attention layers and 16 conditional token embeddings that comprise our $\PromptP\subseteq\mathbb{R}^{768 \times 16}$ space.

\subsection{$\PromptP$ Analysis}



\paragraph{Cross-Attention Analysis}

\begin{figure}
    \centering
    \includegraphics[width=\columnwidth]{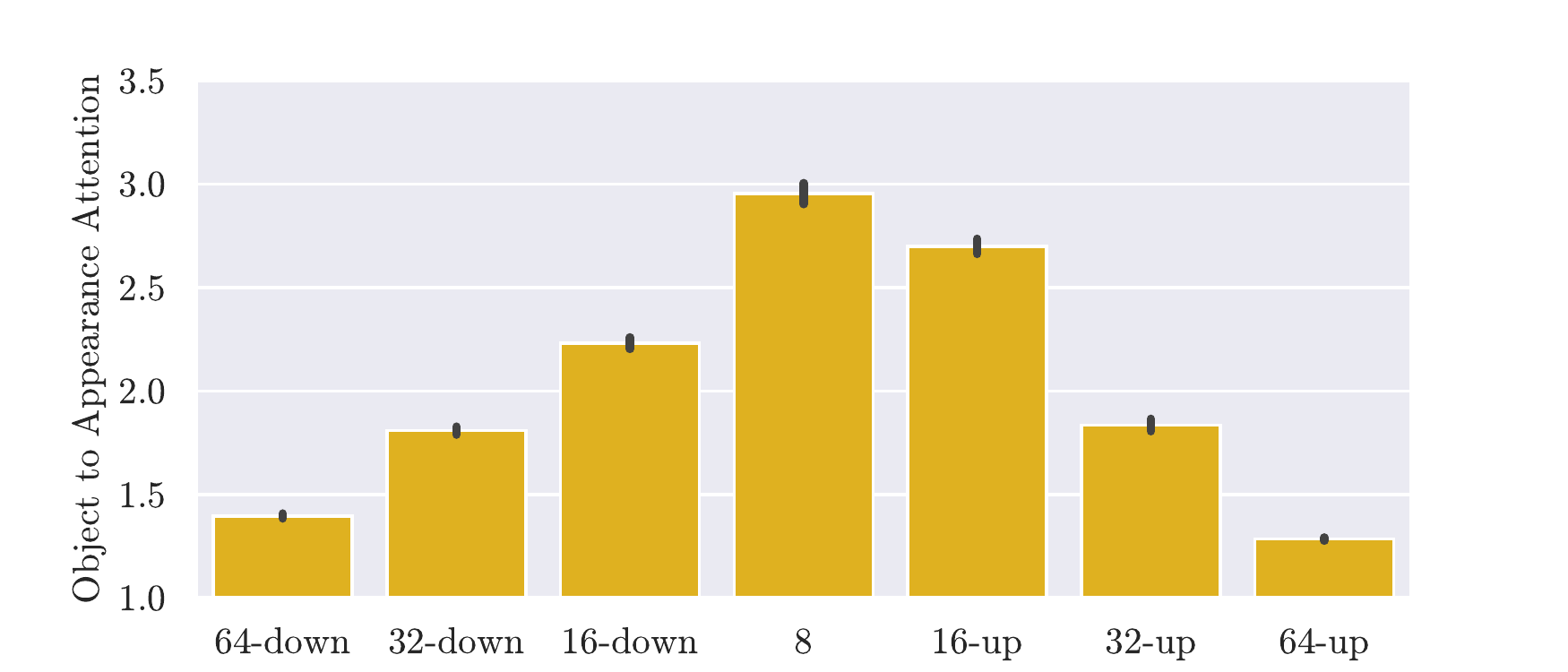}
    \caption{{\bf Object-appearance attention ratio.} Mean ratio of attention features of the object token(s) and appearance token(s), per cross-attention layer. 
    }
    \label{fig:attn_frac}
    \vspace{-10pt}
\end{figure}

We first analyze how the distribution of the cross attention varies across layers. We create a list of 50 objects and 20 appearance adjectives (10 style descriptors and 10 texture descriptors, see supplementary material list). From these lists, we create 2000 ($=50 \times 20 \times 2$) prompts following the patterns \texttt{"appearance object"} and \texttt{"object, appearance"}, and generate 8 images for each prompt using different seeds. We store the cross-attention values for each layer for only the object or appearance token(s), then average over the batch, spatial dimensions, and timesteps to get a ratio of attention on the object token(s) to attention on the appearance token(s). Figure \ref{fig:attn_frac} reports the corresponding ratios. The coarse layers (8, 16) attend proportionally more to the object token and fine layers (32, 64) attend more to the appearance token. This experiment gives us the intuition that coarse layers are more responsible for object shape and structure compared to the fine layers. 


\begin{figure*}[h!]
    \centering
    \includegraphics[width=\textwidth]{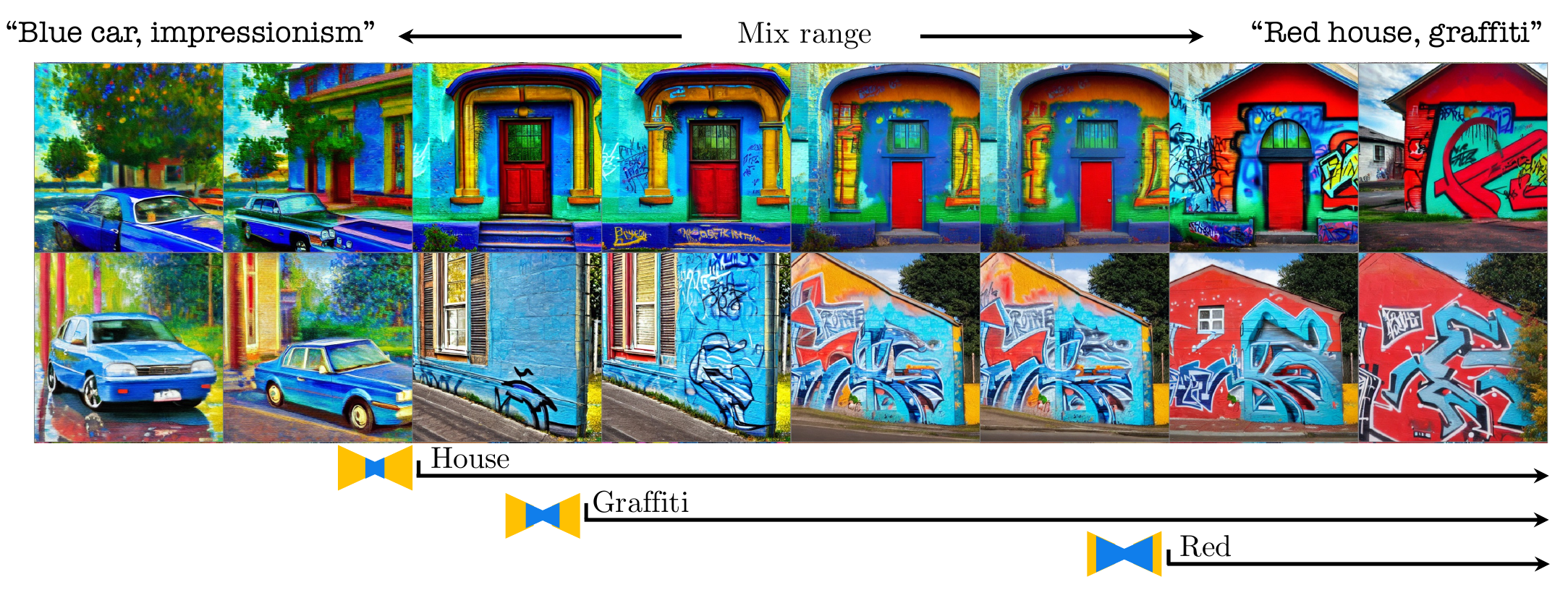}
    \caption{Visualization of mixed conditioning of the U-net cross-attention layers. The rows represent two different starting seeds and the columns represent eight growing subsets of layers, from coarse to fine. We start by conditioning all layers on \texttt{"Blue car, impressionism"} in the left column. As we move right, we gradually condition more layers on \texttt{"Red house, graffiti"}, starting with the innermost coarse layers and then the outer fine layers. Note that \textit{shape} changes (\texttt{"house"}) take place once we condition the coarse layers, but \textit{appearance} (\texttt{"red"}) changes only take place after we condition the fine layers.}
    \label{fig:mix_imgs}
    \vspace{-5pt}
\end{figure*}

\paragraph{Attributes Distribution}
\cq{We further analyze how different cross-attention layers impact different image attributes (shape, color, etc.). To do so, we make use of the CLIP similarity metric \cite{radford2021learning} to quantify the contribution of each layer.}

\begin{figure}
    \centering
    \includegraphics[width=\columnwidth]{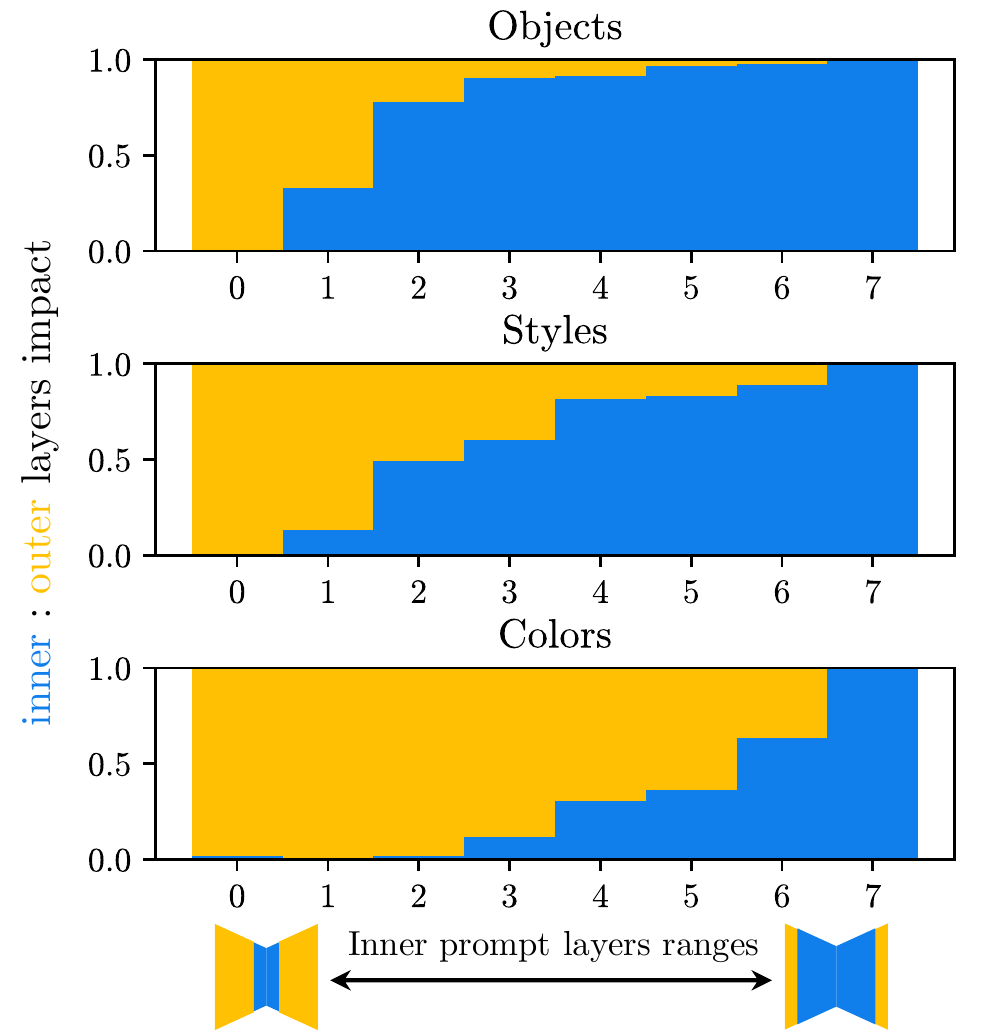}
    \caption{
    Relative CLIP similarities for \texttt{object}, \texttt{color} and \texttt{style} attributes, by subset of U-net layers. {\color{orange}{Orange}} represents the similarity to the first prompt, and {\color{blue}{blue}} represents similarity to the second. As we move from left to right, we gradually grow the subset of layers conditioned with the second prompt from coarse to fine.
    }
    \label{fig:mix_bars}
\end{figure}

First, we divide the $16$ cross-attention layers into $8$ subsets, starting from the the empty set, followed by the middle coarse layer and growing outwards to include the outer fine layers, and finally the full set (see Figure \ref{fig:mix_bars} for a visual explanation and the supplementary for the detailed list).

Next, we take three lists of \texttt{object}, \texttt{color} and \texttt{style} words and randomly generate prompts with the format \texttt{"color object, style"}. For example, \texttt{"green bicycle, oil painting"} or \texttt{"red house, vector art"}. We then randomly sample $64$ pairs of these prompts. For every pair, we condition the aforementioned subset of layers on one prompt, and condition the complement set on the other prompt. We then generate 8 images with fixed seeds for each prompt-pair and subset. 

 Next, we measure the similarity of the output image to each \texttt{object}, \texttt{color} and \texttt{style} attribute with CLIP similarity. This measures the relative contribution of either conditioning prompt.

Figure \ref{fig:mix_imgs} demonstrates this process for a single prompt pair and two image seeds. We start on the left column with all layers conditioned on the first prompt \texttt{"Blue car, impressionism"}. As we move from left to right, we condition more layers from coarse to fine with the other prompt \texttt{"Red house, graffiti"}. Note that even though we already condition some layers on \texttt{"Red house, graffiti"} in the middle column, the house only starts to appear red towards the end when the fine layers are also conditioned on the same prompt.

The results averaged over images and prompt pairs are shown in Figure \ref{fig:mix_bars}. We see that at either extreme, the CLIP similarities are dominated by either prompt (represented as orange or blue). However, like the example in Figure \ref{fig:mix_imgs}, different prompt attributes demonstrate different behaviors in between.
We can see that it is sufficient to condition only the coarse layers for \texttt{object}, while \texttt{color} requires that we condition the full set of layers. \texttt{style} lies somewhere in-between.
Thus, coarse layers determine the \texttt{object} shape and structure of the image, and the fine layers determine the \texttt{color} appearance of the image.  \texttt{style} is a more ambiguous descriptor that involves both shape and texture appearance, so every layer has some contribution towards it. 

\subsection{XTI Evaluation}

Next, we evaluate our proposed XTI and compare our results to the original Textual Inversion (TI)~\cite{gal2022image}.
We use a combined dataset of the TI dataset of 9 concepts, and the dataset from \cite{kumari2022customdiffusion} with 6 concepts. For both datasets, each concept has 4-6 original images.

\cq{We focus on TI as a baseline because it is a model-preserving inversion approach that does not fine tune the model weights. These fine-tuning approaches like DreamBooth~\cite{ruiz2022dreambooth} and Custom Diffusion~\cite{kumari2022customdiffusion} explicitly embed the concept within the model's output domain and thus have excellent reconstruction. However, they have several disadvantages. Firstly, they risk destroying the model's existing prior (catastrophic forgetting). Secondly, they have several orders of magnitude more parameters. Recent work with Low-Rank Adaptation (LoRA)~\cite{lora, lora_diffusion} reduces the number of fine-tuned parameters to a fraction, but this is still about $\sim100$x more than XTI. Lastly, they are difficult to scale to multiple concepts since the fine-tuned parameters for each concept have to be merged.}


\subsubsection{Setup}



We followed the batch size of 8 and performed 5000 optimization steps for Textual Inversion, consistent with the original paper. However, we opted to use a reduced learning rate of 0.005 without scaling for optimization, as opposed to the Latent Diffusion Model from \cite{Rombach_2022_CVPR} used in the original paper. In our experiments, Stable Diffusion with this learning rate worked better. For our proposed XTI, we used the same hyperparameters as for Textual Inversion, except for the number of optimization steps which we reduced to 500, {\bf resulting in significantly faster convergence}. Both Textual Inversion and XTI shared all other hyperparameters, including the placeholder training prompts. On 2$\times$Nvidia A100 GPUs, the whole optimization takes $\sim$15 minutes for XTI compared to $\sim$80 minutes for TI.



\subsubsection{Quantitative Evaluation}

\begin{figure}
    \centering
    \includegraphics[width=\columnwidth]{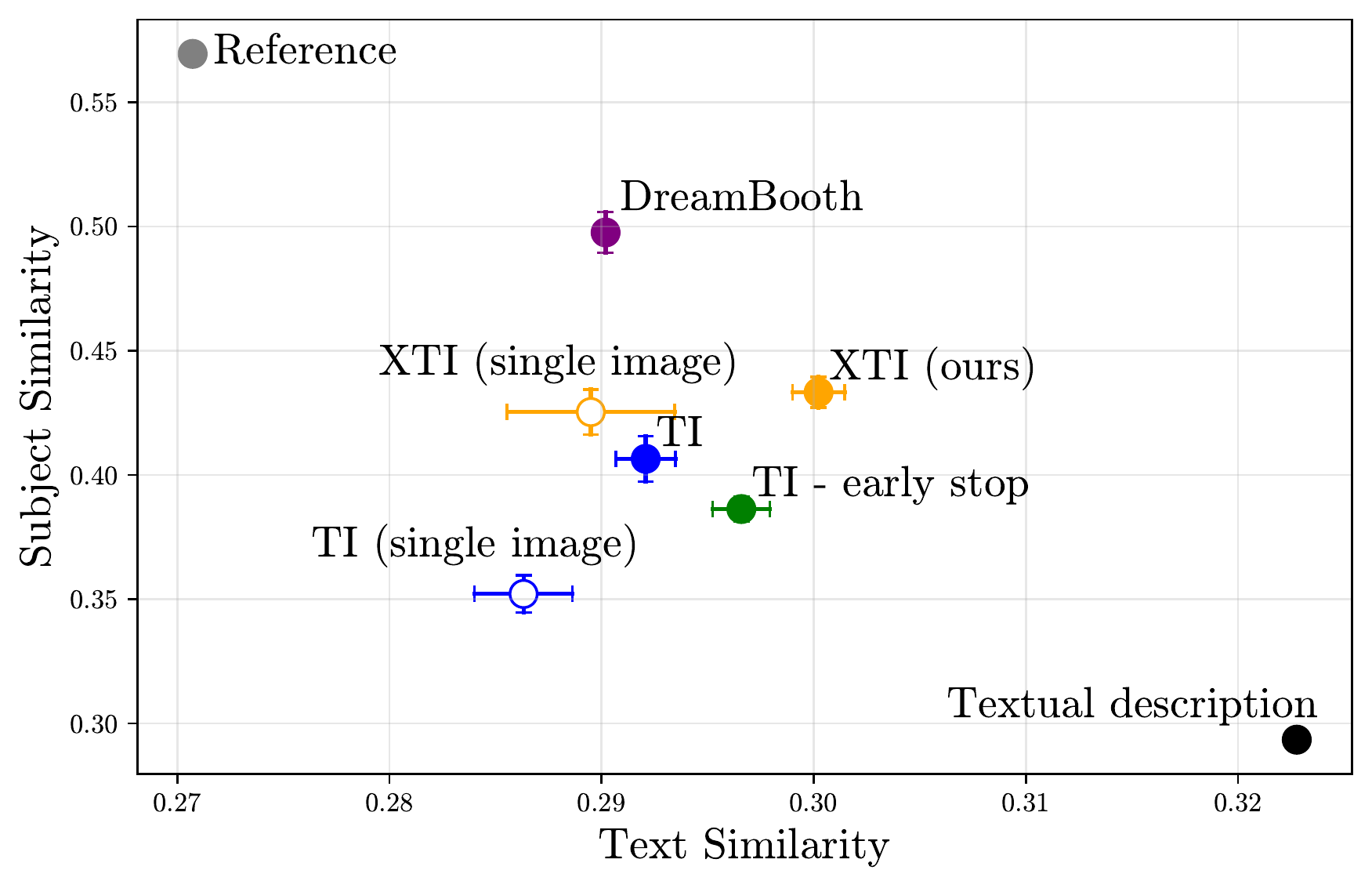}
    \caption{{\bf Comparison of Textual Similarity and Subject Similariy. } Textual Inversion (TI)~\cite{gal2022image}, Extended Textual Inversion (XTI), DreamBooth~\cite{ruiz2022dreambooth}. We also evaluate the metrics for both multi-image and single-image inversion setups. For the latter, a subject is represented by a single image. The ``Reference" label corresponds to images containing the subject images themselves, while the "Textual description" label used the given text description but replaced the explicit subject's description (e.g. "a colorful teapot"). The standard error is visualized in the bars.}
    \label{fig:scatter}
\end{figure}

 Following \cite{gal2022image}, to evaluate the editability quality of the inversions, we use the average cosine similarity between CLIP embeddings of the generated images and the prompts used to generate the images (\textit{Text Similarity}). To measure the distortion of the generated images from the original concept (\textit{Subject Similarity}), we use the average pairwise cosine similarity between ViT-S/16 DINO~\cite{DINO} embeddings of the generated images and the original dataset images. Compared to CLIP which is trained with supervised class labels, \cite{ruiz2022dreambooth} argued that DINO embeddings better capture differences between images of the same class due to its self-supervised training.
 All the methods reported in Figure \ref{fig:scatter} are evaluated over 15 subjects from \cite{gal2022image} and \cite{kumari2022customdiffusion}, each generated with 14 different prompts templates that place the concept in novel context (e.g. \texttt"A photograph of \{\} in the jungles", see Section \ref{sec:metric_prompts} in the supplementary for details). For each test concept and prompt we generated 32 images, making a total of $15 \times 14 \times 32 = 6720$ images. We fix the generation seed across different methods.

In Figure \ref{fig:scatter} we report the evaluation of the proposed Extended Textual Inversion (XTI). Among Textual Inversion \cite{gal2022image}, as for comparison we also include DreamBooth~\cite{ruiz2022dreambooth} which is not a model-preserving method. Notably, XTI outperforms TI at both subject and text similarity despite using 10x fewer training steps.
We also report TI using 500 optimization steps, which is the number of steps we use for XTI. This improves the Text Similarity because fewer number of optimization steps prevents the optimized token embedding from being out of distribution. However, it degrades reconstruction as measured by Subject Similarity. 

We also report the inversion in a data-hungry setup, where the subject is represented with only a single image. Notably, even in this extreme setting the proposed XTI performs better than multi-image TI in terms of subject similarity (see Section \ref{sec:1img} for details).

\begin{figure*}[h]
    \centering
    \includegraphics[width=\textwidth]{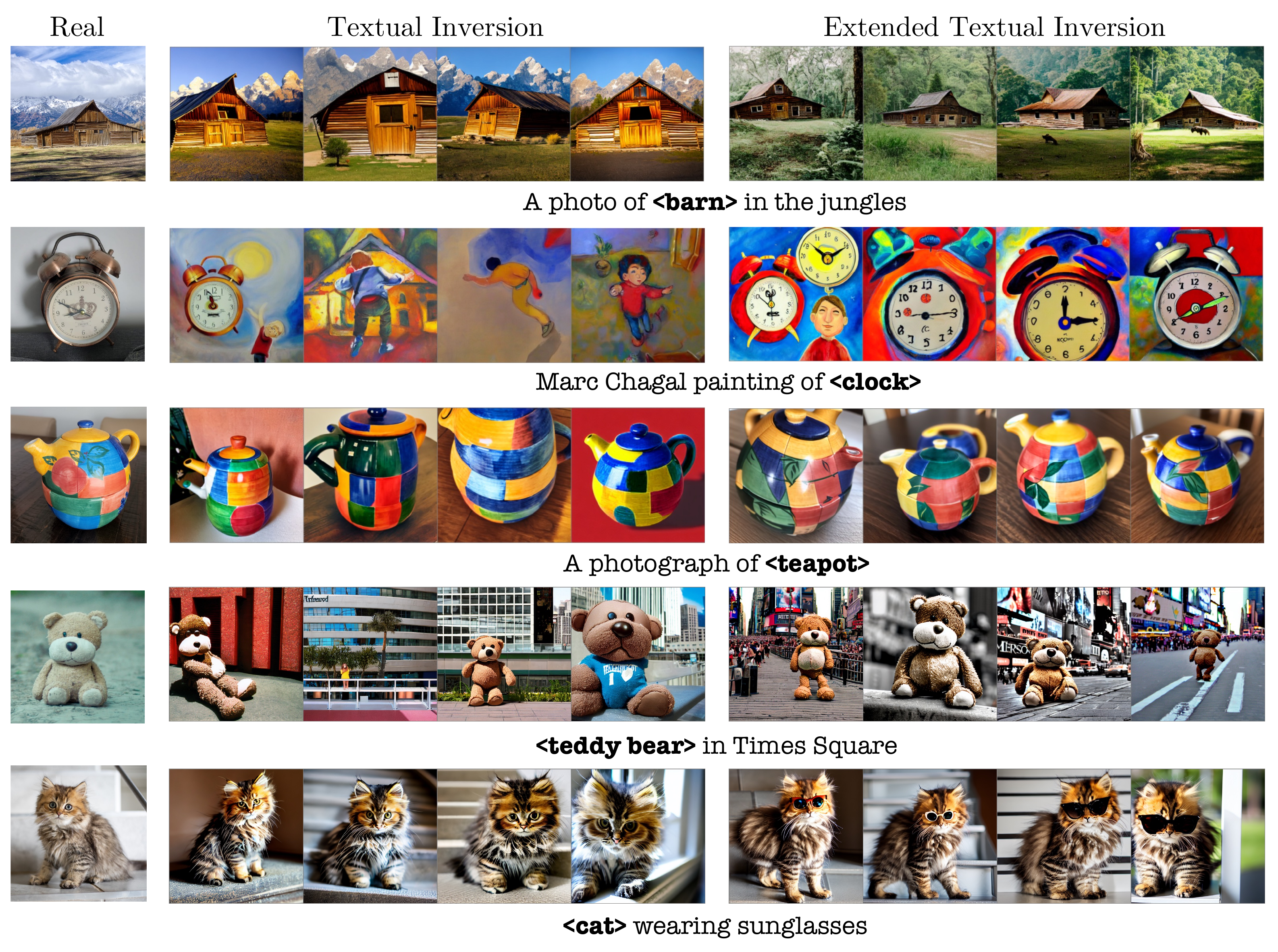}
    \caption{{\bf Textual Inversion (TI) vs. Extended Textual Inversion (XTI).}  \textit{Column 1:} 
    Original concepts. \textit{Column 2:} TI results. \textit{Column 3:} XTI results. It can be seen that XTI exhibits superior subject and prompt fidelity, as corroborated by the results of our user study.}
    \label{fig:xti_samples}
\end{figure*}

\subsubsection{Human Evaluation}

Figure \ref{fig:xti_samples} shows a visual comparison of our XTI approach with the original TI. Our method demonstrates less distortion to the original concept \textit{and} to the target prompt.

To assess the efficacy of our proposed method from a human perspective, we conducted a user study. The study, summarized in Table \ref{table:qualitative_results}, asked participants to evaluate both Textual Inversion (TI) and Extended Textual Inversion (XTI) based on their fidelity to the original subject and the given prompt. The results show a clear preference for XTI for both subject and text fidelity.

\begin{table}[ht]
    \centering
    \begin{tabular}{lrr}
        \textbf{Method} & \textbf{Subject Fidelity} & \textbf{Text Fidelity} \\ \hline
        Textual Inversion & 24\% & 27\% \\ 
        XTI (Ours) & \textbf{76\%} & \textbf{73\%} \\ 
        ~ & ~ & ~ \\ 
    \end{tabular}
    \caption{User study preferences for subject and text fidelity for TI and XTI. See supplementary material for more details.}
    \label{table:qualitative_results}
\end{table}

\subsection{Single Image Inversion}
\label{sec:1img}

\begin{figure*}
    \centering
    \includegraphics[width=\textwidth]{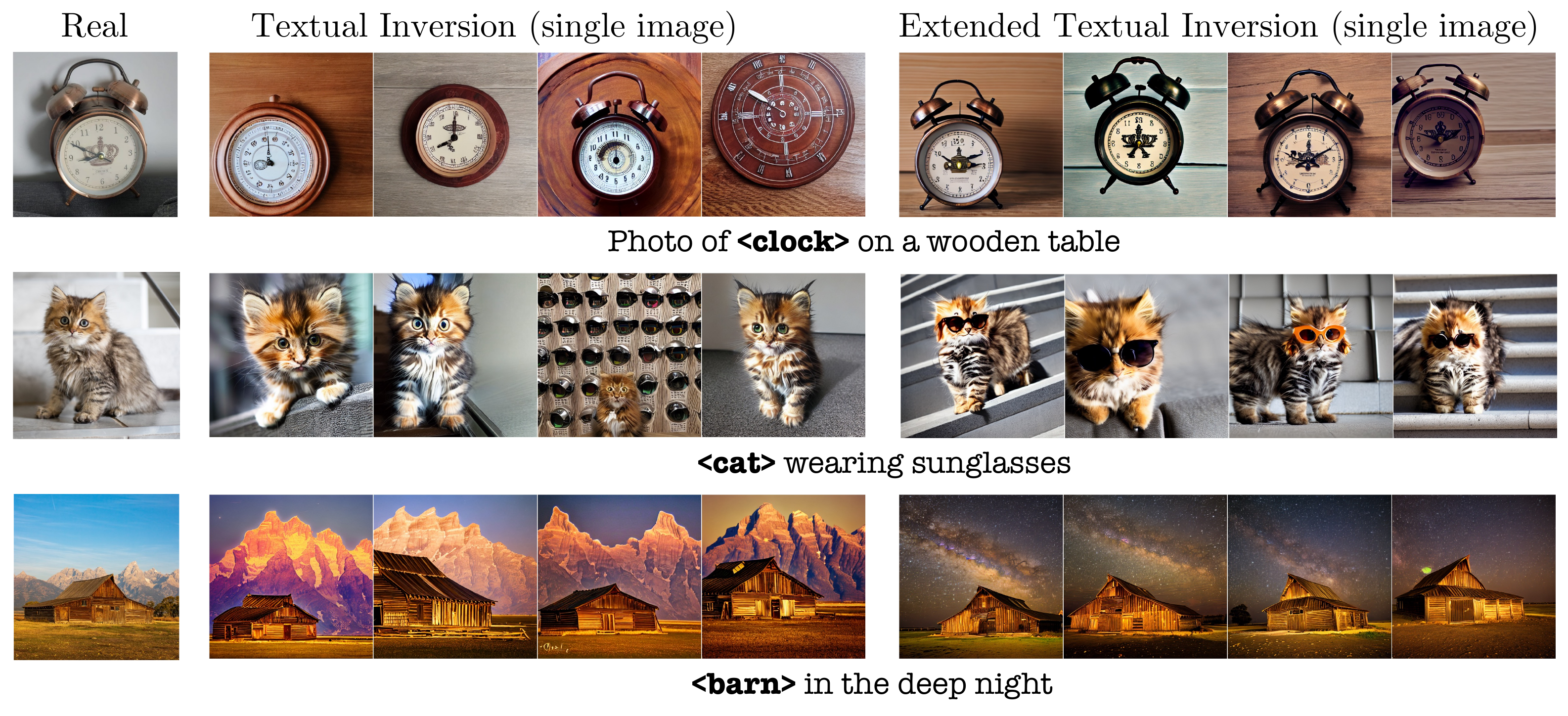}
    \caption{\textbf{Single Image Textual Inversion (TI) vs Single Image Extended Textual Inversion (XTI)}. \textit{Column 1}: Original concepts. \textit{Column 2}: TI results. \textit{Column 3}: XTI results. It can be seen that XTI exhibits superior subject and prompt fidelity and produce plausible results even when trained on a single image.}
    \label{fig:single_image}
\end{figure*}

The Extended Textual Inversion also appears to be very effective in a data-hungry setup, when a target subject is represented with a single image. As for single image training for all the runs we reduce learning rate to $0.001$ to better prevent overfitting. Figure \ref{fig:single_image} provides visual comparison of TI and XTI inversions in the this single image setting. We omit single-image DreamBooth results from Figure~\ref{fig:single_image} and \ref{fig:scatter} due to its comparatively poor performance, namely Text Similarity of 0.25 and Subject Similarity of 0.40. In particular, we found DreamBooth in this single-image setting to be prone to overfitting and difficult to optimize.

\subsection{Embedding Density} \label{sec:kde}

\begin{figure}
    \centering
    \includegraphics[width=\columnwidth]{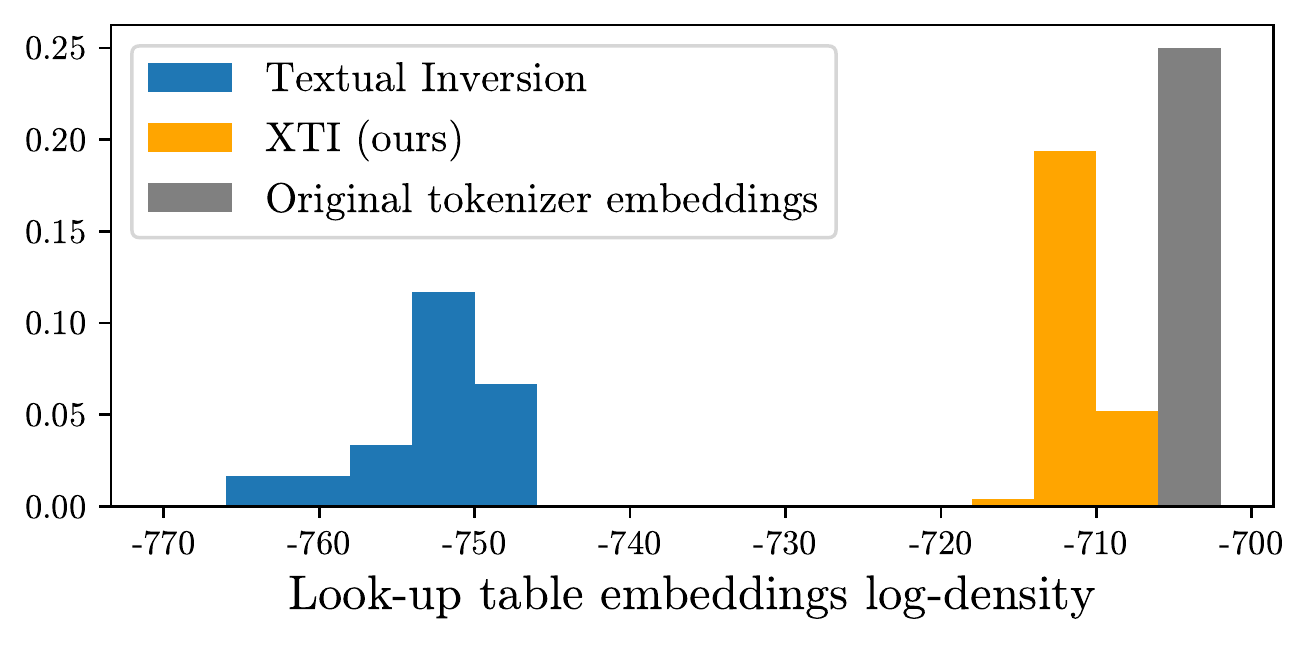}
    \caption{Estimated log-density of the original look-up table token embeddings ({\color{gray}{gray}}), embeddings optimized with textual inversion ({\color{blue}{blue}}), and embeddings optimized with XTI ({\color{orange}{orange}}). Our method demonstrates a more regular representation which is closer to the manifold of natural words.}
    \label{fig:embed_log_density}
\end{figure}

As the textual embeddings inverted with XTI have better editability properties compared to the original TI, this suggests that these tokens are better aligned with the original tokenizer look-up table embedding, \ka{which represents the manifold of natural language embedding}. To quantify this intuition, we evaluate the density of the newly-optimized tokens with respect to the original ``natural" tokens look-up table embeddings. We perform kernel-based density estimation (KDE) in the look-up table tokens embeddings space \cq{independently for each dimension}. Let us define $\mathcal{E}$ to be the set of all original tokens look-up table embeddings, before adding the extra optimized token(s). Assuming that $\mathcal{E}$ is sampled from some continuous distribution, one can define the approximation of its density function at a point $x$ as:

\begin{equation}
\log p_{\mathcal{E}}(x) \approx \frac{1}{|\mathcal{E}|} \sum_{e \in \mathcal{E}} K(x - e), \label{eq:kde}    
\end{equation}
where $K$ is the Gaussian kernel density function \cite{parzen1962estimation, rosenblatt1956remarks}. As for the embeddings optimized with original TI, this quantity always appears to be significantly smaller compared to the densities at the original embeddings $\mathcal{E}$. Figure \ref{fig:embed_log_density} illustrates the original tokens density distribution, and the textual inversion tokens densities. This demonstrates that the proposed approach provides embeddings that are closer to the original distribution, enabling a more natural reconstruction and better editability.

\section{Style Mixing Application}


As we showed earlier, different layers of the denoising U-net are responsible for different aspects of a synthesized image. This allows us to combine the \textit{shape} of one inverted concept with the \textit{appearance} of another inverted concept. We call this Style Mixing. 

Let us consider two independent XTI inversions of two different concepts. We can combine the inversions by passing tokens from different subject to different layers, as illustrated in Figure \ref{fig:teaser_mix}. This mixed conditioning produces an image with a coarse geometry from the first concept, and appearance from the second concept. Formally, we are given two extended prompts: $\{\texttt{p}_n, \dots, \texttt{p}_n\}$, and $\{\texttt{q}_1, \dots, \texttt{q}_n\}$. We form a new extended prompt $\{\texttt{p}_1, \dots, \texttt{p}_k, \texttt{q}_{k+1}, \dots, \texttt{q}_K, \texttt{p}_{K+1}, \dots, \texttt{q}_n\}$ with the separators $1 \leq k < K \leq n$. 


Our observations indicate that the optimization of XTI with an additional density regularization loss term indicated in \ref{eq:kde} enhances its ability to mix objects and styles, without compromising the quality of the inversion output. More details are provided as supplementary material.



\begin{figure}[h!]
    \centering
    \includegraphics[width=\columnwidth]{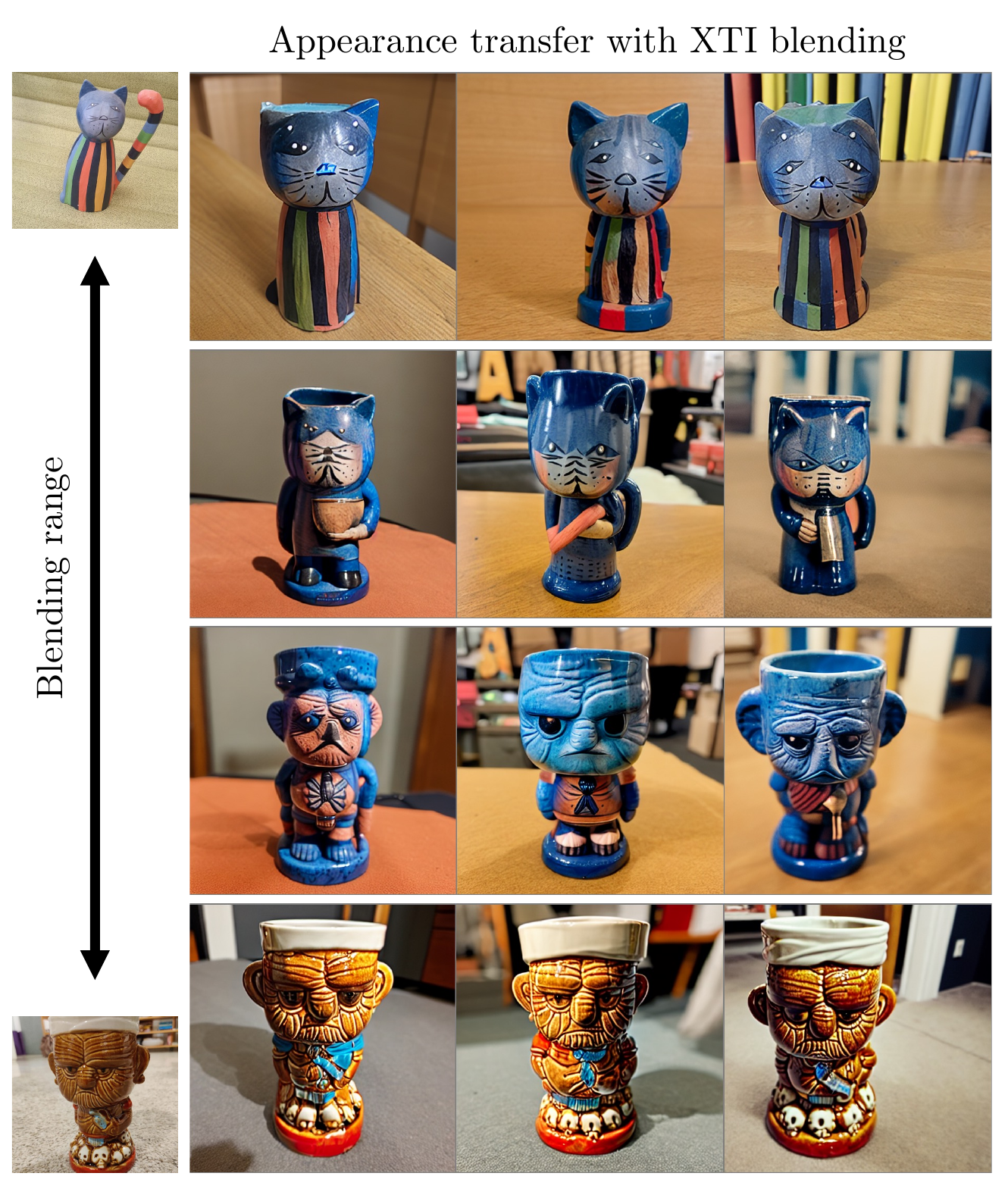}
    \caption{\textbf{Style Mixing in $\PromptP$}. Rows are generated by varying the degree of mixing by adjusting the proportion of layers conditioned on either of the two $\PromptP$ inversions.
    }
    \label{fig:style_mix}
\end{figure}

Figure \ref{fig:style_mix} demonstrates combining the \texttt{"skull mug"} and \texttt{"cat statue"} concepts from \cite{gal2022image}. Different rows of the plot correspond to different blending ranges $k, K$. From top to bottom, we gradually expand it from the middle coarse layer to all the cross-attention layers. This range $(k, K)$ gives the control over the amount of details we want to bring from one inversion to another. By varying $k$ and $K$, we can adjust the contributions of the second subject appearance to the first. 

\begin{figure}[h!]
    \centering
    \includegraphics[width=\columnwidth]{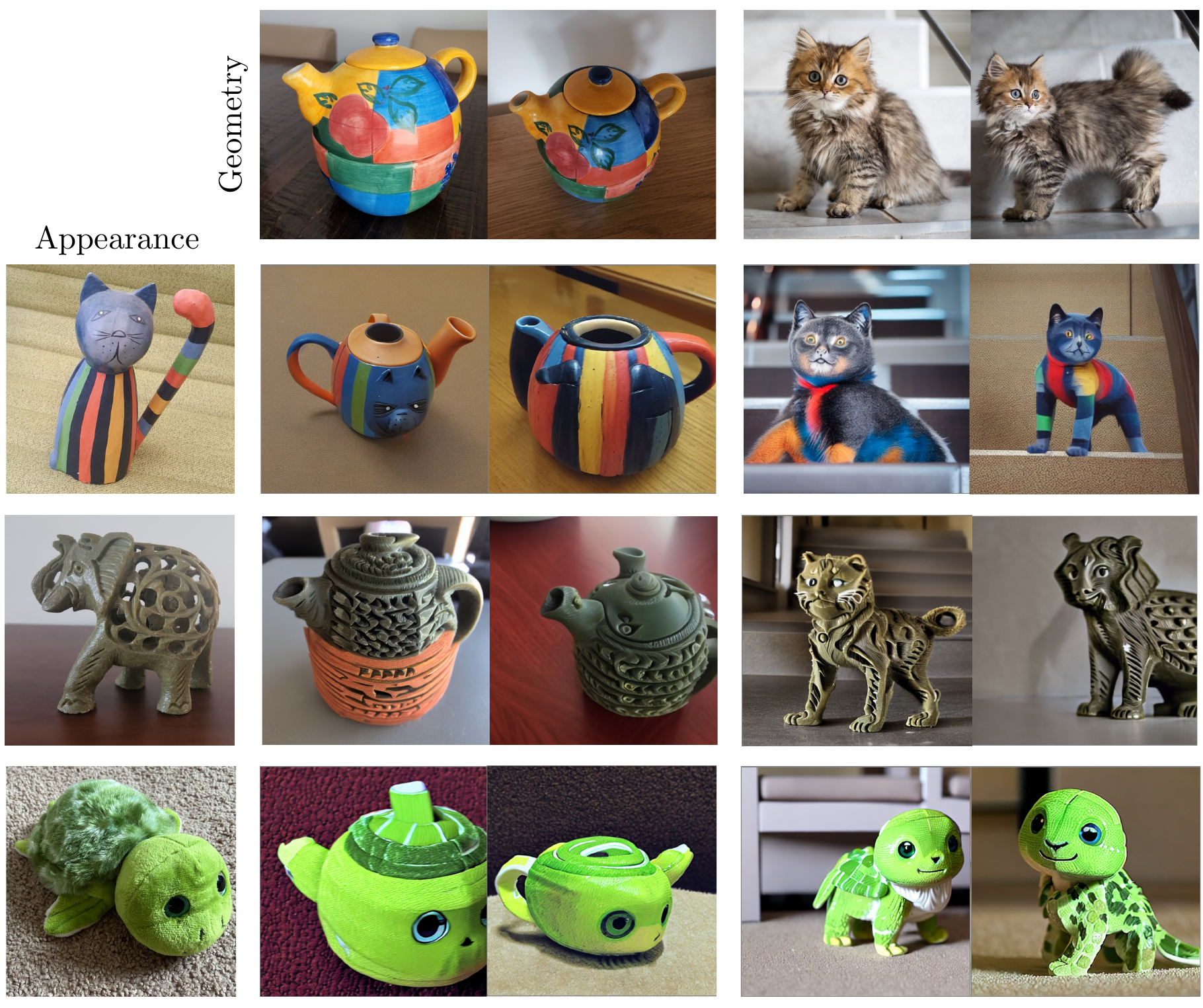}
    \caption{\textbf{More Style Mixing examples}. \textit{Top row}: Shape source concepts. \textit{Left column}: Appearance source concepts.}
    \label{fig:mix_samples}
\end{figure}

Figure \ref{fig:mix_samples} shows a variety of examples generated with this method. Both shape and appearance are inherited remarkably well. More illustrations are provided in supplementary.

Figure \ref{fig:mix_baselines} provides a qualitative comparison between our XTI-based style mixing and baselines of TI~\cite{gal2022image} and DreamBooth~\cite{ruiz2022dreambooth}. The results demonstrate that our approach outperforms the baselines significantly, both in terms of preserving the sources' fidelity and disentangling the attributes.

\begin{figure*}[h!]
    \centering
    \includegraphics[width=0.95\textwidth]{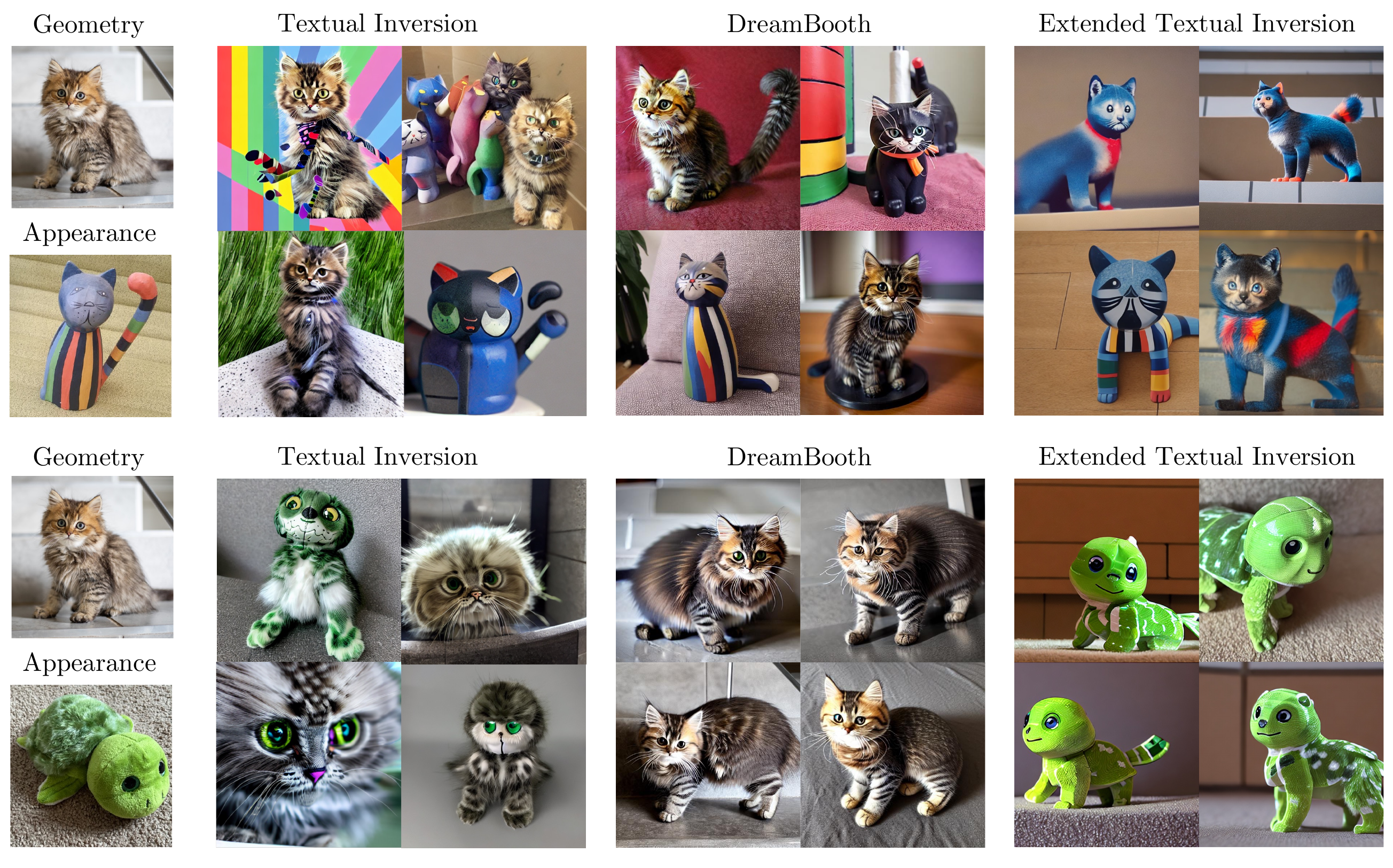}
    \caption{\textbf{Style Mixing comparison.} We compare against Textual Inversion \cite{gal2022image} and Dreambooth \cite{ruiz2022dreambooth} baselines. For TI we independently invert target subject and target appearance, and generate the images with the sentence "\texttt{\textbf{<object>} that looks like \textbf{<appearance>}}". Prompt variations did not make any remarkable improvements. The style source concept was inverted with style prompts (see \cite{gal2022image} for details). DreamBooth was trained with a pair of subjects. Our proposed Extended Textual Inversion clearly outperforms both baselines.}
    \label{fig:mix_baselines}
\end{figure*}
\section{Conclusions, Limitations, and Future work}

We have presented, $\PromptP$, an extended conditional space, which provides 
\cq{increased} expressivity and control.
We have analyzed this space and showed that the denoising U-net demonstrates  disentanglement, where different layers exhibit different sensitivity to \cq{shape or appearance} attributes

The competence of $\PromptP$ is 
\cq{demonstrated} in the Textual Inversion problem. Our Extended Textual Inversion (XTI) is shown to be more accurate, more expressive, more controllable, and significantly faster. Yet surprisingly, we have not observed any reduction in editability.

The performance of XTI, although impressive, is not flawless. Firstly, it does not perfectly reconstruct the concept in the image, and in that respect, it is still inferior to the reconstruction that can be achieved by fine-tuning the model. Secondly, although XTI is significantly faster than TI, it is a rather slow process. Lastly, the disentanglement among the layers of U-net is not perfect, limiting the degree of control that can be achieved through prompt mixing.

An interesting research avenue is to develop encoders to invert one or a few images into $\PromptP$, possibly in the spirit of \cite{gal2023designing}, or to study the impact of applying fine-tuning in conjunction of operating in $\PromptP$. 

\section{Acknowledgement}
We thank Eric Tabellion, Rinon Gal, Miki Rubinstein, Matan Cohen and Jason Baldridge and Yael Pritch for their valuable inputs that helped improve this work.

\newpage

{\small
\bibliographystyle{ieee_fullname}
\bibliography{egbib}

\begin{thebibliography}{10}\itemsep=-1pt

\bibitem{abdal2019image2stylegan}
Rameen Abdal, Yipeng Qin, and Peter Wonka.
\newblock Image2stylegan: How to embed images into the stylegan latent space?
\newblock In {\em Proceedings of the IEEE/CVF International Conference on
  Computer Vision}, pages 4432--4441, 2019.

\bibitem{abdal2020image2stylegan++}
Rameen Abdal, Yipeng Qin, and Peter Wonka.
\newblock Image2stylegan++: How to edit the embedded images?
\newblock In {\em Proceedings of the IEEE/CVF conference on computer vision and
  pattern recognition}, pages 8296--8305, 2020.

\bibitem{alaluf2021hyperstyle}
Yuval Alaluf, Omer Tov, Ron Mokady, Rinon Gal, and Amit~H. Bermano.
\newblock Hyperstyle: Stylegan inversion with hypernetworks for real image
  editing, 2021.

\bibitem{avrahami2022blendedlatent}
Omri Avrahami, Ohad Fried, and Dani Lischinski.
\newblock Blended latent diffusion.
\newblock {\em arXiv preprint arXiv:2206.02779}, 2022.

\bibitem{avrahami2022blended}
Omri Avrahami, Dani Lischinski, and Ohad Fried.
\newblock Blended diffusion for text-driven editing of natural images.
\newblock In {\em Proceedings of the IEEE/CVF Conference on Computer Vision and
  Pattern Recognition}, pages 18208--18218, 2022.

\bibitem{bar2022text2live}
Omer Bar-Tal, Dolev Ofri-Amar, Rafail Fridman, Yoni Kasten, and Tali Dekel.
\newblock Text2live: Text-driven layered image and video editing.
\newblock {\em arXiv preprint arXiv:2204.02491}, 2022.

\bibitem{bau2020units}
David Bau, Jun-Yan Zhu, Hendrik Strobelt, Agata Lapedriza, Bolei Zhou, and
  Antonio Torralba.
\newblock Understanding the role of individual units in a deep neural network.
\newblock {\em Proceedings of the National Academy of Sciences}, 2020.

\bibitem{bermano2022state}
Amit~H Bermano, Rinon Gal, Yuval Alaluf, Ron Mokady, Yotam Nitzan, Omer Tov,
  Oren Patashnik, and Daniel Cohen-Or.
\newblock State-of-the-art in the architecture, methods and applications of
  stylegan.
\newblock In {\em Computer Graphics Forum}, volume~41, pages 591--611. Wiley
  Online Library, 2022.

\bibitem{brooks2022instructpix2pix}
Tim Brooks, Aleksander Holynski, and Alexei~A. Efros.
\newblock Instructpix2pix: Learning to follow image editing instructions.
\newblock In {\em CVPR}, 2023.

\bibitem{DINO}
Mathilde Caron, Hugo Touvron, Ishan Misra, Herv{\'{e}} J{\'{e}}gou, Julien
  Mairal, Piotr Bojanowski, and Armand Joulin.
\newblock Emerging properties in self-supervised vision transformers.
\newblock {\em CoRR}, abs/2104.14294, 2021.

\bibitem{chang2023muse}
Huiwen Chang, Han Zhang, Jarred Barber, AJ Maschinot, Jose Lezama, Lu Jiang,
  Ming-Hsuan Yang, Kevin Murphy, William~T Freeman, Michael Rubinstein, et~al.
\newblock Muse: Text-to-image generation via masked generative transformers.
\newblock {\em arXiv preprint arXiv:2301.00704}, 2023.

\bibitem{couairon2022diffedit}
Guillaume Couairon, Jakob Verbeek, Holger Schwenk, and Matthieu Cord.
\newblock Diffedit: Diffusion-based semantic image editing with mask guidance.
\newblock {\em arXiv preprint arXiv:2210.11427}, 2022.

\bibitem{creswell2018inverting}
Antonia Creswell and Anil~Anthony Bharath.
\newblock Inverting the generator of a generative adversarial network.
\newblock {\em IEEE transactions on neural networks and learning systems},
  30(7):1967--1974, 2018.

\bibitem{crowson2022vqgan}
Katherine Crowson, Stella Biderman, Daniel Kornis, Dashiell Stander, Eric
  Hallahan, Louis Castricato, and Edward Raff.
\newblock Vqgan-clip: Open domain image generation and editing with natural
  language guidance.
\newblock {\em arXiv preprint arXiv:2204.08583}, 2022.

\bibitem{gal2022image}
Rinon Gal, Yuval Alaluf, Yuval Atzmon, Or Patashnik, Amit~H Bermano, Gal
  Chechik, and Daniel Cohen-Or.
\newblock An image is worth one word: Personalizing text-to-image generation
  using textual inversion.
\newblock {\em arXiv preprint arXiv:2208.01618}, 2022.

\bibitem{gal2023designing}
Rinon Gal, Moab Arar, Yuval Atzmon, Amit~H Bermano, Gal Chechik, and Daniel
  Cohen-Or.
\newblock Designing an encoder for fast personalization of text-to-image
  models.
\newblock {\em arXiv preprint arXiv:2302.12228}, 2023.

\bibitem{hertz2022prompt}
Amir Hertz, Ron Mokady, Jay Tenenbaum, Kfir Aberman, Yael Pritch, and Daniel
  Cohen-Or.
\newblock Prompt-to-prompt image editing with cross attention control.
\newblock {\em arXiv preprint arXiv:2208.01626}, 2022.

\bibitem{ho2020denoising}
Jonathan Ho, Ajay Jain, and Pieter Abbeel.
\newblock Denoising diffusion probabilistic models.
\newblock {\em Advances in Neural Information Processing Systems},
  33:6840--6851, 2020.

\bibitem{ho2021classifier}
Jonathan Ho and Tim Salimans.
\newblock Classifier-free diffusion guidance.
\newblock In {\em NeurIPS 2021 Workshop on Deep Generative Models and
  Downstream Applications}, 2021.

\bibitem{lora}
Edward~J. Hu, Yelong Shen, Phillip Wallis, Zeyuan Allen-Zhu, Yuanzhi Li, Shean
  Wang, Lu Wang, and Weizhu Chen.
\newblock Lora: Low-rank adaptation of large language models, 2021.

\bibitem{Karras2020ada}
Tero Karras, Miika Aittala, Janne Hellsten, Samuli Laine, Jaakko Lehtinen, and
  Timo Aila.
\newblock Training generative adversarial networks with limited data.
\newblock In {\em Proc. NeurIPS}, 2020.

\bibitem{karras2019style}
Tero Karras, Samuli Laine, and Timo Aila.
\newblock A style-based generator architecture for generative adversarial
  networks.
\newblock In {\em Proceedings of the IEEE conference on computer vision and
  pattern recognition}, pages 4401--4410, 2019.

\bibitem{kawar2022imagic}
Bahjat Kawar, Shiran Zada, Oran Lang, Omer Tov, Huiwen Chang, Tali Dekel, Inbar
  Mosseri, and Michal Irani.
\newblock Imagic: Text-based real image editing with diffusion models.
\newblock {\em arXiv preprint arXiv:2210.09276}, 2022.

\bibitem{Kawar2022ImagicTR}
Bahjat Kawar, Shiran Zada, Oran Lang, Omer Tov, Hui-Tang Chang, Tali Dekel,
  Inbar Mosseri, and Michal Irani.
\newblock Imagic: Text-based real image editing with diffusion models.
\newblock {\em ArXiv}, abs/2210.09276, 2022.

\bibitem{kim2021diffusionclip}
Gwanghyun Kim and Jong~Chul Ye.
\newblock Diffusionclip: Text-guided image manipulation using diffusion models.
\newblock {\em arXiv preprint arXiv:2110.02711}, 2021.

\bibitem{kumari2022customdiffusion}
Nupur Kumari, Bingliang Zhang, Richard Zhang, Eli Shechtman, and Jun-Yan Zhu.
\newblock Multi-concept customization of text-to-image diffusion.
\newblock {\em CVPR}, 2023.

\bibitem{kwon2021clipstyler}
Gihyun Kwon and Jong~Chul Ye.
\newblock Clipstyler: Image style transfer with a single text condition.
\newblock {\em arXiv preprint arXiv:2112.00374}, 2021.

\bibitem{lin2022magic3d}
Chen-Hsuan Lin, Jun Gao, Luming Tang, Towaki Takikawa, Xiaohui Zeng, Xun Huang,
  Karsten Kreis, Sanja Fidler, Ming-Yu Liu, and Tsung-Yi Lin.
\newblock Magic3d: High-resolution text-to-3d content creation.
\newblock {\em arXiv preprint arXiv:2211.10440}, 2022.

\bibitem{lipton2017precise}
Zachary~C Lipton and Subarna Tripathi.
\newblock Precise recovery of latent vectors from generative adversarial
  networks.
\newblock {\em arXiv preprint arXiv:1702.04782}, 2017.

\bibitem{pndm}
Luping Liu, Yi Ren, Zhijie Lin, and Zhou Zhao.
\newblock Pseudo numerical methods for diffusion models on manifolds, 2022.

\bibitem{meng2021sdedit}
Chenlin Meng, Yang Song, Jiaming Song, Jiajun Wu, Jun-Yan Zhu, and Stefano
  Ermon.
\newblock Sdedit: Image synthesis and editing with stochastic differential
  equations.
\newblock {\em arXiv preprint arXiv:2108.01073}, 2021.

\bibitem{mokady2022null}
Ron Mokady, Amir Hertz, Kfir Aberman, Yael Pritch, and Daniel Cohen-Or.
\newblock Null-text inversion for editing real images using guided diffusion
  models.
\newblock {\em arXiv preprint arXiv:2211.09794}, 2022.

\bibitem{nichol2021glide}
Alex Nichol, Prafulla Dhariwal, Aditya Ramesh, Pranav Shyam, Pamela Mishkin,
  Bob McGrew, Ilya Sutskever, and Mark Chen.
\newblock Glide: Towards photorealistic image generation and editing with
  text-guided diffusion models.
\newblock {\em arXiv preprint arXiv:2112.10741}, 2021.

\bibitem{nitzan2022mystyle}
Yotam Nitzan, Kfir Aberman, Qiurui He, Orly Liba, Michal Yarom, Yossi
  Gandelsman, Inbar Mosseri, Yael Pritch, and Daniel Cohen-Or.
\newblock Mystyle: A personalized generative prior.
\newblock {\em ACM Transactions on Graphics (TOG)}, 41(6):1--10, 2022.

\bibitem{parmar2023zero}
Gaurav Parmar, Krishna~Kumar Singh, Richard Zhang, Yijun Li, Jingwan Lu, and
  Jun-Yan Zhu.
\newblock Zero-shot image-to-image translation.
\newblock {\em arXiv preprint arXiv:2302.03027}, 2023.

\bibitem{parzen1962estimation}
Emanuel Parzen.
\newblock On estimation of a probability density function and mode.
\newblock {\em The annals of mathematical statistics}, 33(3):1065--1076, 1962.

\bibitem{patashnik2021styleclip}
Or Patashnik, Zongze Wu, Eli Shechtman, Daniel Cohen-Or, and Dani Lischinski.
\newblock Styleclip: Text-driven manipulation of stylegan imagery.
\newblock {\em arXiv preprint arXiv:2103.17249}, 2021.

\bibitem{radford2021learning}
Alec Radford, Jong~Wook Kim, Chris Hallacy, Aditya Ramesh, Gabriel Goh,
  Sandhini Agarwal, Girish Sastry, Amanda Askell, Pamela Mishkin, Jack Clark,
  et~al.
\newblock Learning transferable visual models from natural language
  supervision.
\newblock {\em arXiv preprint arXiv:2103.00020}, 2021.

\bibitem{ramesh2022hierarchical}
Aditya Ramesh, Prafulla Dhariwal, Alex Nichol, Casey Chu, and Mark Chen.
\newblock Hierarchical text-conditional image generation with clip latents.
\newblock {\em arXiv preprint arXiv:2204.06125}, 2022.

\bibitem{richardson2020encoding}
Elad Richardson, Yuval Alaluf, Or Patashnik, Yotam Nitzan, Yaniv Azar, Stav
  Shapiro, and Daniel Cohen-Or.
\newblock Encoding in style: a stylegan encoder for image-to-image translation.
\newblock {\em arXiv preprint arXiv:2008.00951}, 2020.

\bibitem{roich2021pivotal}
Daniel Roich, Ron Mokady, Amit~H. Bermano, and Daniel Cohen-Or.
\newblock Pivotal tuning for latent-based editing of real images.
\newblock {\em ACM Transactions on Graphics (TOG)}, 2022.

\bibitem{rombach2021highresolution}
Robin Rombach, Andreas Blattmann, Dominik Lorenz, Patrick Esser, and Björn
  Ommer.
\newblock High-resolution image synthesis with latent diffusion models, 2021.

\bibitem{Rombach_2022_CVPR}
Robin Rombach, Andreas Blattmann, Dominik Lorenz, Patrick Esser, and Bj\"orn
  Ommer.
\newblock High-resolution image synthesis with latent diffusion models.
\newblock In {\em Proceedings of the IEEE/CVF Conference on Computer Vision and
  Pattern Recognition (CVPR)}, pages 10684--10695, June 2022.

\bibitem{rosenblatt1956remarks}
Murray Rosenblatt.
\newblock Remarks on some nonparametric estimates of a density function.
\newblock {\em The annals of mathematical statistics}, pages 832--837, 1956.

\bibitem{ruiz2022dreambooth}
Nataniel Ruiz, Yuanzhen Li, Varun Jampani, Yael Pritch, Michael Rubinstein, and
  Kfir Aberman.
\newblock Dreambooth: Fine tuning text-to-image diffusion models for
  subject-driven generation.
\newblock {\em arXiv preprint arXiv:2208.12242}, 2022.

\bibitem{lora_diffusion}
Simo Ryu.
\newblock Low-rank adaptation for fast text-to-image diffusion fine-tuning.
\newblock \url{https://github.com/cloneofsimo/lora}, 2022.

\bibitem{saharia2022photorealistic}
Chitwan Saharia, William Chan, Saurabh Saxena, Lala Li, Jay Whang, Emily
  Denton, Seyed Kamyar~Seyed Ghasemipour, Burcu~Karagol Ayan, S~Sara Mahdavi,
  Rapha~Gontijo Lopes, Tim Salimans, Tim Salimans, Jonathan Ho, David~J Fleet,
  and Mohammad Norouzi.
\newblock Photorealistic text-to-image diffusion models with deep language
  understanding.
\newblock {\em arXiv preprint arXiv:2205.11487}, 2022.

\bibitem{sohl2015deep}
Jascha Sohl-Dickstein, Eric Weiss, Niru Maheswaranathan, and Surya Ganguli.
\newblock Deep unsupervised learning using nonequilibrium thermodynamics.
\newblock In {\em International Conference on Machine Learning}, pages
  2256--2265. PMLR, 2015.

\bibitem{song2020denoising}
Jiaming Song, Chenlin Meng, and Stefano Ermon.
\newblock Denoising diffusion implicit models.
\newblock In {\em International Conference on Learning Representations}, 2020.

\bibitem{song2019generative}
Yang Song and Stefano Ermon.
\newblock Generative modeling by estimating gradients of the data distribution.
\newblock {\em Advances in Neural Information Processing Systems}, 32, 2019.

\bibitem{tov2021designing}
Omer Tov, Yuval Alaluf, Yotam Nitzan, Or Patashnik, and Daniel Cohen-Or.
\newblock Designing an encoder for stylegan image manipulation.
\newblock {\em arXiv preprint arXiv:2102.02766}, 2021.

\bibitem{pnpDiffusion2022}
Narek Tumanyan, Michal Geyer, Shai Bagon, and Tali Dekel.
\newblock Plug-and-play diffusion features for text-driven image-to-image
  translation.
\newblock {\em arXiv preprint arXiv:2211.12572}, 2022.

\bibitem{valevski2022unitune}
Dani Valevski, Matan Kalman, Yossi Matias, and Yaniv Leviathan.
\newblock Unitune: Text-driven image editing by fine tuning an image generation
  model on a single image.
\newblock {\em arXiv preprint arXiv:2210.09477}, 2022.

\bibitem{voynovrpgan}
Andrey Voynov and Artem Babenko.
\newblock Rpgan: random paths as a latent space for gan interpretability.

\bibitem{wang2022imagen}
Su Wang, Chitwan Saharia, Ceslee Montgomery, Jordi Pont-Tuset, Shai Noy,
  Stefano Pellegrini, Yasumasa Onoe, Sarah Laszlo, David~J Fleet, Radu Soricut,
  et~al.
\newblock Imagen editor and editbench: Advancing and evaluating text-guided
  image inpainting.
\newblock {\em arXiv preprint arXiv:2212.06909}, 2022.

\bibitem{Wang2021HighFidelityGI}
Tengfei Wang, Yong Zhang, Yanbo Fan, Jue Wang, and Qifeng Chen.
\newblock High-fidelity gan inversion for image attribute editing.
\newblock {\em ArXiv}, abs/2109.06590, 2021.

\bibitem{wu2021stylespace}
Zongze Wu, Dani Lischinski, and Eli Shechtman.
\newblock Stylespace analysis: Disentangled controls for stylegan image
  generation.
\newblock In {\em Proceedings of the IEEE/CVF Conference on Computer Vision and
  Pattern Recognition}, pages 12863--12872, 2021.

\bibitem{xia2021gan}
Weihao Xia, Yulun Zhang, Yujiu Yang, Jing-Hao Xue, Bolei Zhou, and Ming-Hsuan
  Yang.
\newblock Gan inversion: A survey, 2021.

\bibitem{yeh2017semantic}
Raymond~A. Yeh, Chen Chen, Teck~Yian Lim, Alexander~G. Schwing, Mark
  Hasegawa-Johnson, and Minh~N. Do.
\newblock Semantic image inpainting with deep generative models, 2017.

\bibitem{zeiler2014visualizing}
Matthew~D Zeiler and Rob Fergus.
\newblock Visualizing and understanding convolutional networks.
\newblock In {\em Computer Vision--ECCV 2014: 13th European Conference, Zurich,
  Switzerland, September 6-12, 2014, Proceedings, Part I 13}, pages 818--833.
  Springer, 2014.

\bibitem{zhu2016generative}
Jun-Yan Zhu, Philipp Kr{\"a}henb{\"u}hl, Eli Shechtman, and Alexei~A Efros.
\newblock Generative visual manipulation on the natural image manifold.
\newblock In {\em European conference on computer vision}, pages 597--613.
  Springer, 2016.

\end{thebibliography}
}

\newpage
{\centering \Large{Supplementary}}

\subsection{Regularization}

When applying style mixing, we discovered that optimizing XTI with an additional density regularization loss term (Equation 2) improves the mixing capability while maintaining the overall quality of the inversion. To achieve this, we use an extra loss term $-\lambda \cdot \sum_{i=1}^l \log p_{\mathcal{E}}(e_i)$, where $\lambda = 0.002$ serves as a small regularization scale. This loss term encourages the newly added look-up table embedding to be even more regular, causing the optimized token distribution from Figure \ref{fig:embed_log_density} to shift closer to the original token distribution. We contend that this is particularly advantageous for the mixture application because in this scenario, generation is conditioned by two different XTI tokens, making it crucial to have them interact naturally, i.e., with the two tokens lying closer to the natural language manifold.

However, applying this regularization term to the original TI for subject recontextualization leads to a degradation in subject similarity of the inversion. Even with a small $\lambda$ scaling factor, this regularization enforces significant simplification in the inversion process, leaving the reconstructed token with limited freedom and expressivity. Meanwhile, although the drop in quality for XTI is minimal when the regularization is added, we do not use it by default for the recontextualization task for XTI because it would increase its complexity (another hyperparameter) and convergence rate.


\begin{figure*}
    \centering
    \includegraphics[width=0.875\textwidth]{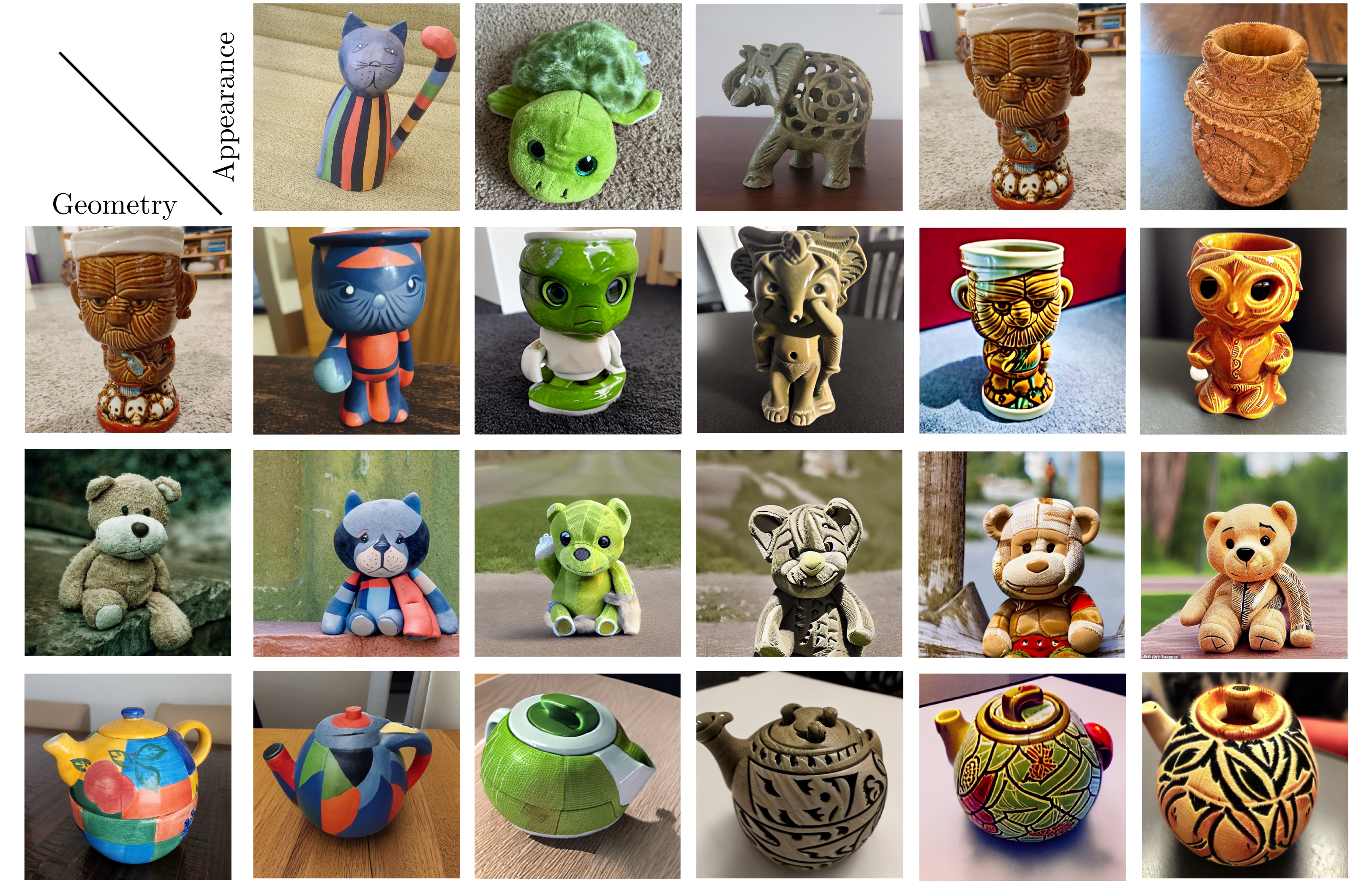}
    \caption{\textbf{Style mixing examples}. \textit{Top row}: Style source subjects, \textit{First column}: Geometry source subjects. The geometry subject's tokens are passed to the three layers in the range \texttt{(16,~'down',~1) - (16,~'up',~0)}, while all the rest are conditioned on the appearance subject's token.}
    \label{fig:mix_inner}
\end{figure*}
\begin{figure*}
    \centering
    \includegraphics[width=0.875\textwidth]{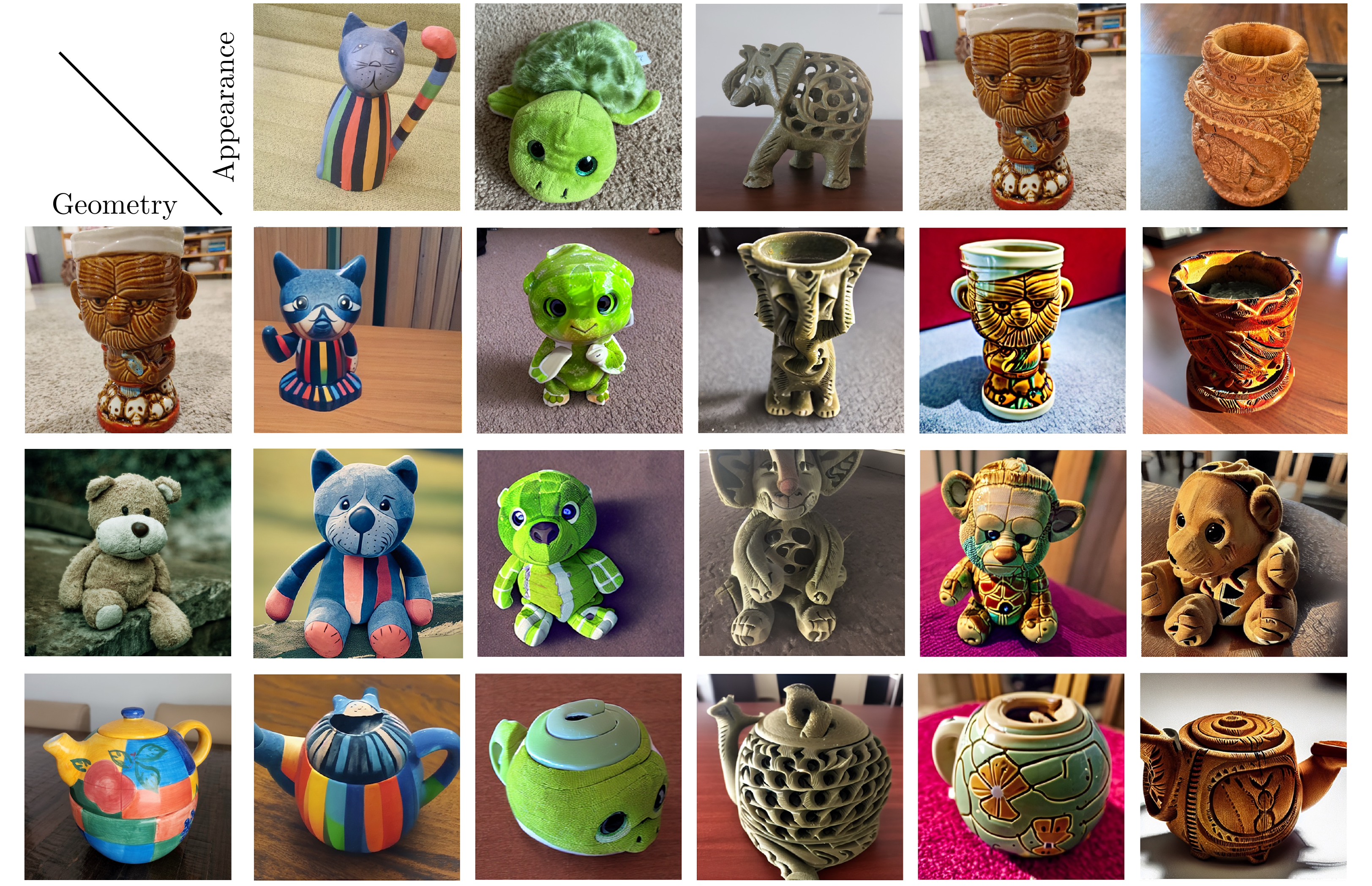}
    \caption{\textbf{Style mixing with more appearance layers}. \textit{Top row}: Style source subjects, \textit{First column}: Geometry source subjects. Here the geometry subject's tokens are passed only to two layers \texttt{(8,~'down',~0)} and \texttt{(16,~'up',~0)}. Thus this emphasizes the appearance subject's token more, resulting in a more dominant appearance compared to the previous setup in Figure~\ref{fig:mix_inner}.}
    \label{fig:mix_inner_2}
\end{figure*}


\section{Further Results and Details}

Figures \ref{fig:mix_inner} and \ref{fig:mix_inner_2} provide more examples of geometry and style mixing. In Figure \ref{fig:mix_inner} objects prompts are passed to a wider range of layers compared to Figure \ref{fig:mix_inner_2}, enforcing higher source geometry alignment.

\begin{figure*}[h!]
    \centering
    \includegraphics[width=0.95\textwidth]{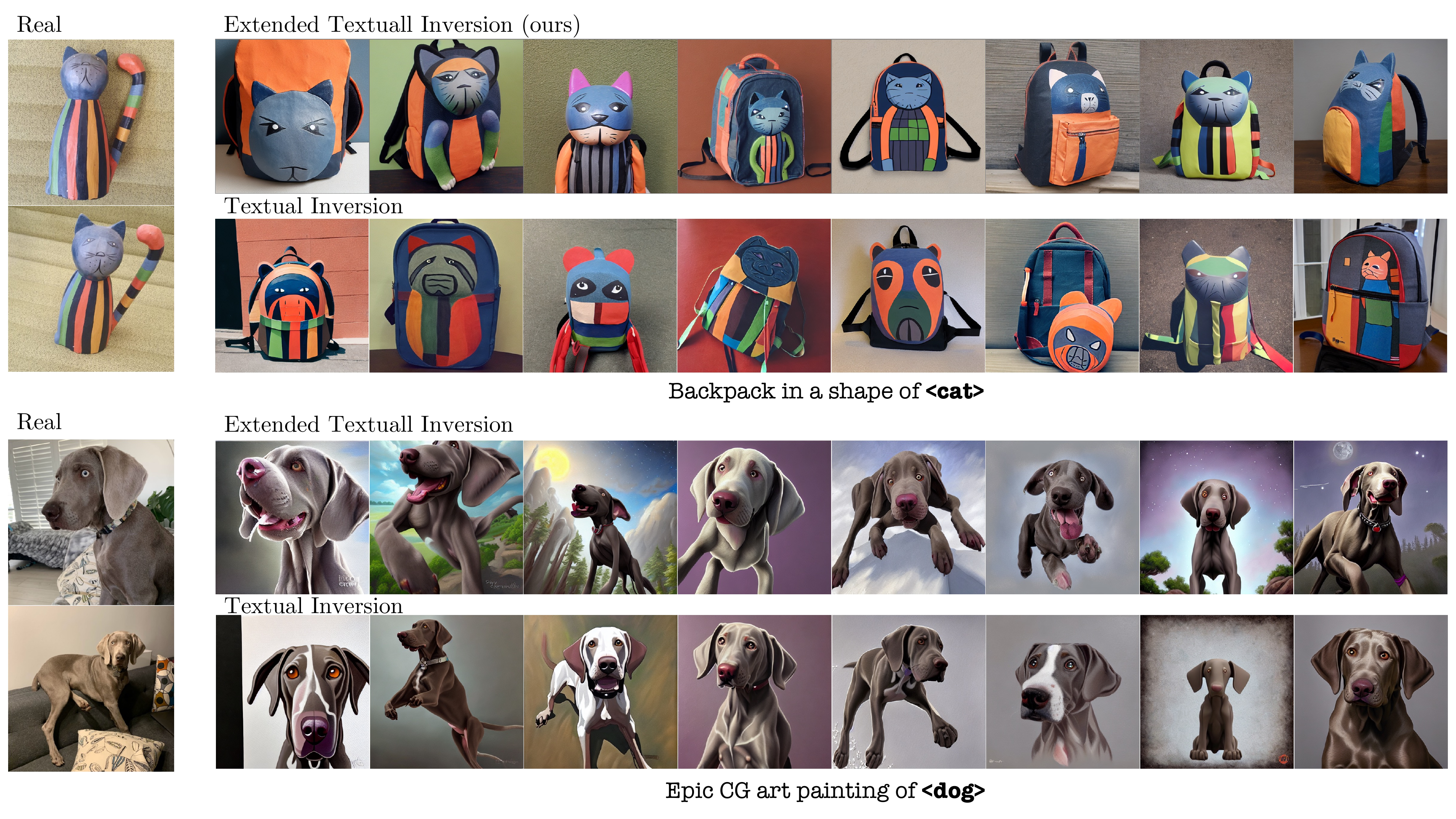}
    \caption{Samples generated with Textual Inversion (TI) and the proposed Extended Textual Inversion (XTI). XTI has better text alignment while providing more accurate subject reconstruction.}
    \label{fig:samples_all}
\end{figure*}

Figure \ref{fig:samples_all} provides more uncurated examples generated with Textual Inversion and Extended Textual Inversion.

As for inference we use the PNDM scheduler \cite{pndm} with 50 denoising steps and a classifier-free guidance \cite{ho2021classifier} scale of 7.5. 

We implemented DreamBooth with a learning rate of 5e-6, a batch size of 4, and conducted 400 finetuning steps per-subject, using Stable Diffusion. Our optimization focused solely on the U-net weights and did not involve prior preservation. 

\subsection{Coarse/Fine Layers Split}
We provide the details of the layer subsets we used in the shape-style mixing experiments in Figures \ref{fig:attn_frac} and \ref{fig:mix_bars}. We name the cross-attention layers of Stable Diffusion U-net as follows, in the order they appear in the U-net: \\
\texttt{(64,~'down',~0), (64,~'down',~1), (32,~'down',~0), (32,~'down',~1), (16,~'down',~0), (16,~'down',~1), (8,~'down',~0), (16,~'up',~0), (16,~'up',~1), (16,~'up', 2), (32,~'up',~0), (32,~'up',~1), (32,~'up', 2), (64,~'up',~0), (64,~'up',~1), (64,~'up', 2)}. \\
\\
The first number represents the spatial resolution, \texttt{'down' / 'up'} represents whether the layer is on the downward (contracting) or upward (expansive) part of the U-net, and the third number indicates the index among the cross-attention layers of the same resolution and direction.

In Figure \ref{fig:mix_bars} we use the following growing sequence of cross-attention layer subsets: \\
0: Empty set \\
1: Layer \texttt{(8,~'down', 0)} only \\
2: \texttt{(16,~'down',~1)~-~(8,~'down',~0)} \\
3: \texttt{(16,~'down',~1)~-~(16,~'up',~0)} \\
4: \texttt{(16,~'down',~0)~-~(16,~'up',~0)} \\
5: \texttt{(16,~'down',~0)~-~(16,~'up',~1)} \\
6: \texttt{(16,~'down',~0)~-~(16,~'up',~2)} \\
7: \texttt{(64,~'down',~0)~-~(64,~'up',~2)} \\
The ranges listed above are inclusive.
\\

In Figures 2 and 11 in the main text we condition the layers \texttt{(8,~'down',~0), (16,~'up',~0)} on the target shape textual embeddings, and the other layers on the target style textual embeddings. In Figure 3 in the main text we provide the target shape textual embeddings to layers \texttt{(16,~'down',~1) - (16,~'up',~0)}.

\subsection{Text Prompts}

\subsubsection{Cross-Attention Analysis (Fig. 5)}

We used the following lists of objects and appearances for generating the \texttt{"appearance object"} and \texttt{"object, appearance"} prompts in Figure 5:

Objects (50): \texttt{"dog", "cat", "tree", "chair", "book", "phone", "car", "bike", "lamp", "table", "flower", "desk", "computer", "pen", "pencil", "lamp", "television", "picture", "mirror", "shoe", "boot", "sandals", "house", "building", "street", "park", "river", "ocean", "lake", "mountain", "chair", "couch", "armchair", "bookcase", "rug", "lampshade", "fan", "conditioner", "heater", "door", "window", "bed", "pillow", "blanket", "curtains", "kitchen", "refrigerator", "stove", "oven", "microwave"}

Appearances (20): \texttt{"fuzzy", "shiny", "bright", "fluffy", "sparkly", "dull", "smooth", "rough", "jagged", "striped", "painting", "retro", "vintage", "modern", "bohemian", "industrial", "rustic", "classic", "contemporary", "futuristic"}
\\

\subsubsection{Image Attributes Analysis (Fig. \ref{fig:mix_bars})}
For Figure \ref{fig:mix_bars}, we used the following object, color and style words to generate the prompts:

Objects (13): \texttt{"chair", "dog", "book", "elephant", "guitar", "pillow", "rabbit", "umbrella", "yacht", "house", "cube", "sphere", 'car'}

Colors (11): \texttt{"black", "blue", "brown", "gray", "green", "orange", "pink", "purple", "red", "white", "yellow"}

Style descriptions (7): \texttt{"watercolor", "oil painting", "vector art", "pop art style", "3D rendering", "impressionism picture", "graffiti"}

\subsubsection{Text Similarity Metric Prompts}

\label{sec:metric_prompts}
For Text Similarity evaluation we use the following 14 prompts:

\texttt{"A photograph of <token>", "A photo of <token> in the jungles", "A photo of <token> on a beach", "Aquarelle painting of <token>", "Oil painting of <token>", "Marc Chagall painting of <token>", "Sketch drawing of <token>", "Night photograph of <token>", "Professional studio photograph of <token>", "3d rendering of <token>", "Fantasy CG art painting of <token>", "A statue of <token>", "A photograph of two <token> on a table", "App icon of <token>"}. 

Here $\texttt{<token>}$ represents the placeholder for XTI inversion tokens to be replaced with the corresponding textual description in Table~\ref{table:detailed_captions}.

\begin{table*}[ht]
\centering
\begin{tabular}{ll}
\vspace{5pt}
\textbf{Original Dataset} & \textbf{Text Description} \\
\texttt{elephant} & \texttt{a statue of an elephant} \\
\texttt{cat\_statue} & \texttt{a statue of a cat} \\
\texttt{colorful\_teapot} & \texttt{a colorful teapot} \\
\texttt{clock} & \texttt{an alarm clock} \\
\texttt{mug\_skulls} & \texttt{a cup with a mummy} \\
\texttt{physics\_mug} & \texttt{a black cup with math equations} \\
\texttt{red\_teapot} & \texttt{a red teapot} \\
\texttt{round\_bird} & \texttt{a round bird sculpture} \\
\texttt{thin\_bird} & \texttt{a sculpture of a thin bird} \\
\texttt{barn} & \texttt{an old wooden barn} \\
\texttt{cat} & \texttt{a kitten} \\
\texttt{dog} & \texttt{a grey dog} \\
\texttt{teddybear} & \texttt{a teddy bear} \\
\texttt{tortoise\_plushy} & \texttt{a tortoise plush} \\
\texttt{wooden\_pot} & \texttt{an artistic wooden pot} \\
\end{tabular}
\vspace{10pt}
\caption{Detailed text descriptions for each dataset. The first 9 correspond to the datasets provided in \cite{gal2022image}, and the remaining 6 correspond to the datasets provided in \cite{kumari2022customdiffusion}.}
\label{table:detailed_captions}
\end{table*}



\begin{figure*}
    \centering
    \includegraphics[width=\textwidth]{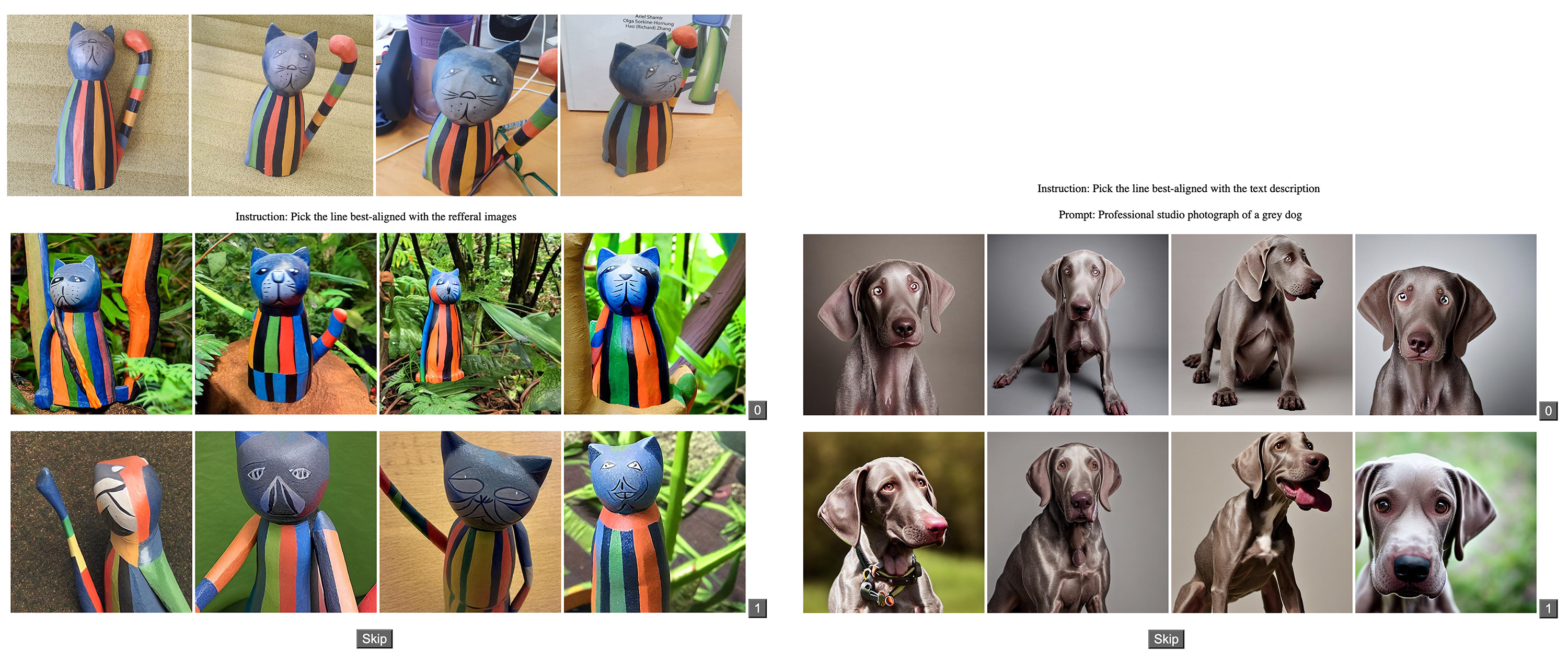}
    \caption{Human labeling interface. On the \textit{left} we depict a sample task to evaluate subject similarity, and on the \textit{right} the task to evaluate text similarity. The comparing methods raws are always shuffled. Both methods use the same random seed.}
    \label{fig:my_label}
\end{figure*}

\end{document}